# Aerial Image Object Detection With Vision Transformer Detector (ViTDet)

*Liya Wang, Alex Tien*

The MITRE Corporation, McLean, VA, 22102, United States

## ABSTRACT

The past few years have seen an increased interest in aerial image object detection due to its critical value to large-scale geoscientific research like environmental studies, urban planning, and intelligence monitoring. However, the task is very challenging due to the bird's-eye view perspective, complex backgrounds, large and various image sizes, different appearances of objects, and the scarcity of well-annotated datasets. Recent advances in computer vision have shown promise tackling the challenge. Specifically, Vision Transformer Detector (ViTDet) was proposed to extract multi-scale features for object detection. The empirical study shows that ViTDet's simple design achieves good performance on natural scene images and can be easily embedded into any detector architecture. To date, ViTDet's potential benefit to challenging aerial image object detection has not been explored. Therefore, in our study, 25 experiments were carried out to evaluate the effectiveness of ViTDet for aerial image object detection on three well-known datasets: Airbus Aircraft, RarePlanes, and Dataset of Object DeTection in Aerial images (DOTA). Our results show that ViTDet can consistently outperform its convolutional neural network counterparts on horizontal bounding box (HBB) object detection by a large margin (up to 17% on average precision) and that it achieves the competitive performance for oriented bounding box (OBB) object detection. Our results also establish a baseline for future research.

## 1. INTRODUCTION

Aerial image object detection has been a vibrant research topic for its essential role in large-scale geoscientific research like environmental science, ecology, agricultural studies, wildfire monitoring, urban planning, intelligence monitoring, and emergency rescue. However, the task is very challenging due to the bird's-eye view perspective, complex backgrounds, large and various image sizes, various appearances of objects, and the scarcity of well-annotated datasets [1]. In addition, the objects in aerial images are often arbitrarily oriented. For that reason, instead of using common horizontal bounding boxes (HBBs) (Fig. 1a), oriented bounding boxes (OBBs) (Fig. 1b) have been alternatively used to avoid mismatching between bounding boxes and corresponding objects [1].

In the past few years, deep learning techniques have dominated the object detection domain for their effective feature learning capability. Fig. 2 shows the milestones of deep learning algorithms in object detection since 2014. The green and orange rectangles in Fig. 2 highlight one-stage and two-stage methods for HBB object detection, respectively. Two-stage methods have two separate processes, region proposal and detection proposal, while in one-stage methods these two processes are combined. In general, two-stage methods have better performance than one-stage methods at the expense of computational workload. The purple rectangle in Fig. 2 shows the algorithms designed specifically for OBB object detection; the yellow rectangle calls out the important deep learning frameworks for building object detection algorithms, where ResNet [2], Vision Transformer (ViT) [3], and Swin-T [4] are commonly used backbones for feature extraction; in particular, Vision Transformer Detector (ViTDet) [5] was a newly proposed backbone for object detection; feature pyramid network (FPN) [6] is often used as neck after backbone for feature fusion.

Feature learning is always essential in any computer vision (CV) machine learning methods. Following the advent of ViTs [3] in 2021, an exciting self-supervised learning (SSL) method, Masked Autoencoder (MAE) [7], was proposed to learn effective visual representation features. MAE adopts Masked Image Modeling (MIM) technique and tries to infer masked image patches from unmasked ones. To date, MAE has attracted unprecedented attention because of its superior performance over its supervised learning and contrastive learning counterparts. The encouraging success of MAE has inspired a wide range of applications in the areas of video, audio, medical images, earth observation, multimodal, point cloud, 3D mesh data, reinforcement learning, and graphs (see Table 1 for the summary). It is noticeable that several efforts have been devoted to object detection, including ViTDet [5], a new backbone designed especially for object detection with support of MAE pretrained ViT.

ViTDet was particularly designed to enhance the effectiveness of ViT backbone on object detection problems. Although MAE pretrained ViTs are effective for image classification tasks, they are less effective for object detection, which usually requires multi-scale features. ViT is a plain, non-hierarchical architecture that maintains a single-scale feature map throughout, which indicates that ViT åbackbone is not sufficient for object detection tasks, especially when dealing with multi-scale objects and high-resolution images [8]. To deal with the deficiency, ViTDet was invented to extract multi-scale features for object detection. ViTDet has demonstrated its superior performance on natural scene image (e.g., COCO [9]) object detection [5], and its simple design can also make it embeddable in any detector architecture.

To the authors' knowledge, no research has ever adopted ViTDet and examined its performance for aerial image object detection. Therefore, this research aims to evaluate and gain insight on the potential benefits of ViTDet for both aerial image HBB and OBB object detection. The remainder of the paper is organized as follows: Section II describes related work. Section III discusses the aerial image datasets used for performance evaluation. Section IV gives the details of the implementation platforms for carrying out the experiments. The results are presented in Section V. Section VI is the conclusion.

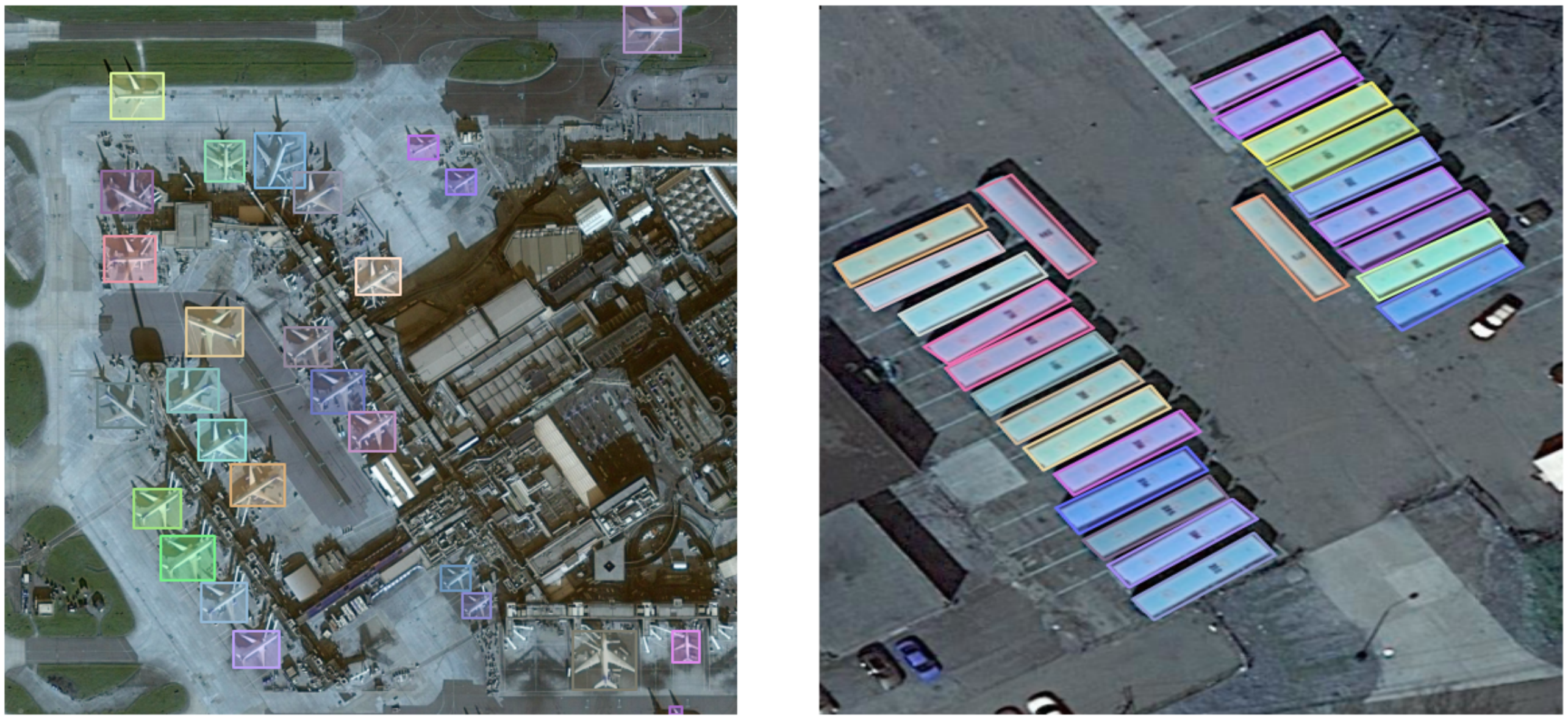

**(a). HBB** **(b). OBB**

**Fig. 1 Illustration of different bounding box types.**

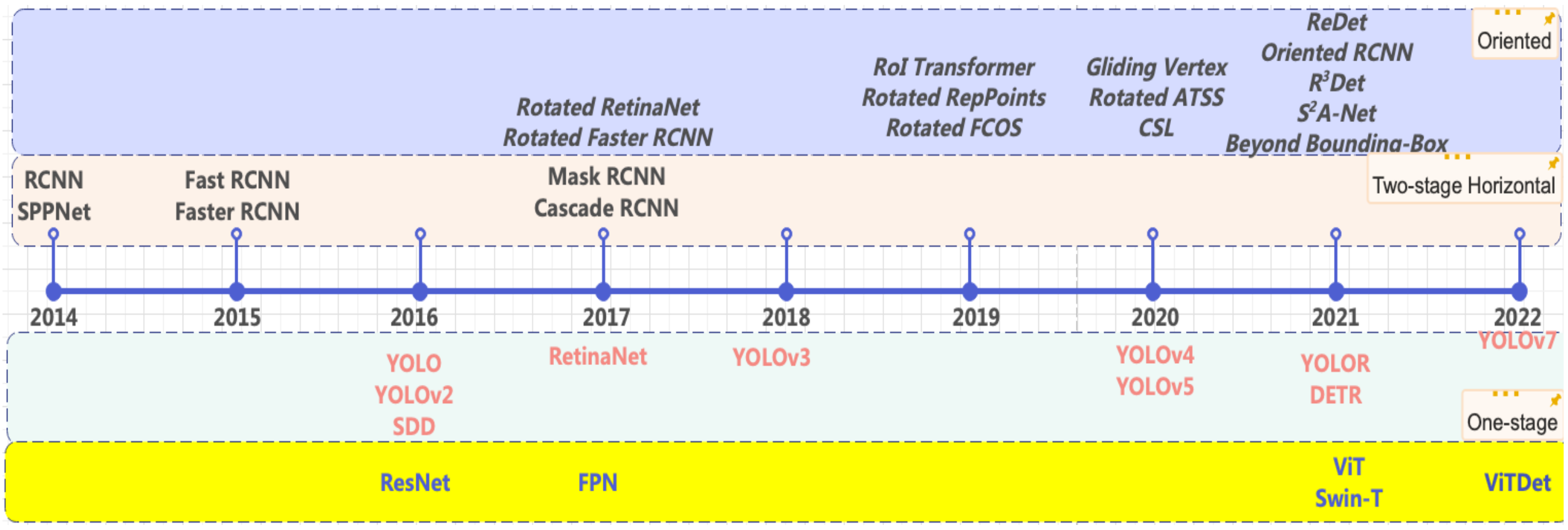

**Fig. 2 Milestones of deep learning algorithms in object detection.**

**Table 1 Summary of latest MIM research**

| Domain | Sub-Domain | Research Papers |
|---|---|---|
| Vision | Image | BEiT v1 [10], v2 [11], MAE [7], SimMIM [12], ADIOS [13], AMT [14], AttMask [15], Beyond-Masking [16], BootMAE [17], CAE [18], CAN [19], ConvMAE [20], Contrastive MAE [21], ContrastMask [22], dBOT [23], DMAE [24], Denoising MAE [25], GreenMAE [26], iBOT [27], LoMaR [28], LS-MAE [29], MaskAlign [30], MaskDistill [31], MaskFeat [32], MaskTune [33], MetaMask [34], MFM [35], MILAN [36], MixMask [37], MixMIM [38], MRA [39], MSN [40], MST [41], MultiMAE [42], MVP [43], RC-MAE [44], SDMAE [45], SemMAE [46], SdAE [47], SupMAE [48], U-MAE [49], UM-MAE [50] |
| | Video | AdaMAE [51], Bevt [52], $MAM^2$ [53], MAR [54], MaskViT [55], $M^3$Video [56], MCVD [57], MotionMAE [58], OmniMAE [59], Spatial-Temporal [60], SSVH [61], VideoMAE [62], Vimpac [63], VRL [64] |
| | Medical Image | DAMA [65], GCMAE [66], SD-MAE [67], SMIT [68] |
| | Satellite Image | SatMAE [69] |
| | Image Classification | MUST [70] |
| | Object Detection | imTED [71], Mask DINO [72], ObjMAE [73], PACMAC [74], ViTDet [5] |
| | Segmentation | kMaX-DeepLab [75], Mask-CLIP [76], MaskDistill [77], Mask Transfiner [78], MOVE [79], NameMask [80] |
| | Image Generation | DiffEdit [81], MAGE [82], MaskGIT [83], Divide-and-Revise [84] |
| | Face Recognition | FaceMAE [85], FFR-Net [86], MFR [87] |
| | Text Recognition | MaskOCR [88] |
| Multimodal | Vision-Language | Data2vec [89], M3AE [90], MAMO [91], MaskCLIP [92], Masked V+L [93], $M^3$AE [94], MLIM [95], ViCHA [96], VL-BEiT [97], VLC [98], VIOLET [99], VLMAE [100] |
| | Audio-Language | CAV-MAE [101] |
| Others | Audio | Audio-MAE [102], Group-MAE [103], MAE-AST [104], MSM [105], M2D [106] |
| | Anomaly Detection | MAEDAY [107], SSMCTB [108], ST-MAE [109] |
| | Graph | MGAP [110], GMAE [111], GMAE-AS [112], GraphMAE [113], HGMAE [114], MGAE [115], MaskGAE [116] |
| | Point Cloud | Point-Bert [117], Point-MAE [118], Point-M2AE [119], Mask-Point [120], Masked [121], Voxel-MAE [122] |
| | Skeleton | SimMC [123] |
| | Depth Estimation | Depth Refinement [124], FM-Net [125] |
| | Reinforcement Learning | MLR [126], Motor Control, Visual Control [127] |
| | 3D Mesh Data | MeshMAE [128] |
| | Adversarial Attack | MAD [129] |
| | Miscellaneous | D-MAE [130], MAEEG [131], MGD [132], Extra-MAE , MADE [133], MaskDP [134], i2i [135], Lifetime Prediction [136], MET [137], MIL [138], Robot Training [139], Time Series [140] |
| Survey | | MIM Survey [141] |
| Theory | | CL vs MIM [142], Contextual Representation Learning[143] , Data Scaling [144], EVA [145], i-MAE [146], Revealing MIM [147], Understanding MAE [148], Understanding MIM [149], Understanding DR [150], |
| Architecture | | Deeper vs Wider [151], Masked BNN [152], ViT Back to CNN [153], ConvNeXt V2 [154] |

## 2. RELATED RESEARCH

### 2.1 Backbones

In CV deep learning methods, backbones are commonly used to extract discriminative object feature representation, and they have been a driving force for rapid object detection performance improvement [155]. Popular backbones for object detection are ResNet [2], ResNeXt [156], and Swin-T [4] because of their deep hierarchical architectures, which can produce the needed multi-scale features. Backbone pretraining is usually carried out on ImageNet-1k [157] with either supervised learning or SSL methods like contrastive learning or MAE, which will be presented next.

### 2.2 MAE

MAE is an asymmetric autoencoder that uses ViTs in both its encoder and decoder, and the size of decoder is smaller than the encoder, as illustrated in Fig. 3. It directly infers masked patches from the unmasked ones with a simple loss of mean squared error (MSE). To save computation, the encoder works on only the unmasked patches; in contrast, the decoder works on both masked and unmasked patches trying to predict the original images. The masking ratio can be set up to 75%, which is considerably higher than that in BERT (typically 15%) [158] or earlier MIM methods (20% to 50%) [10], [159]. MAE's ablation study also points out that a high masking ratio is good for fine-tuning and linear probing [7]. With those meticulous designs, MAE is three times (or more) faster than Bidirectional Encoder representation from Image Transformers (BEiT) [10] while achieving superior performance [7].

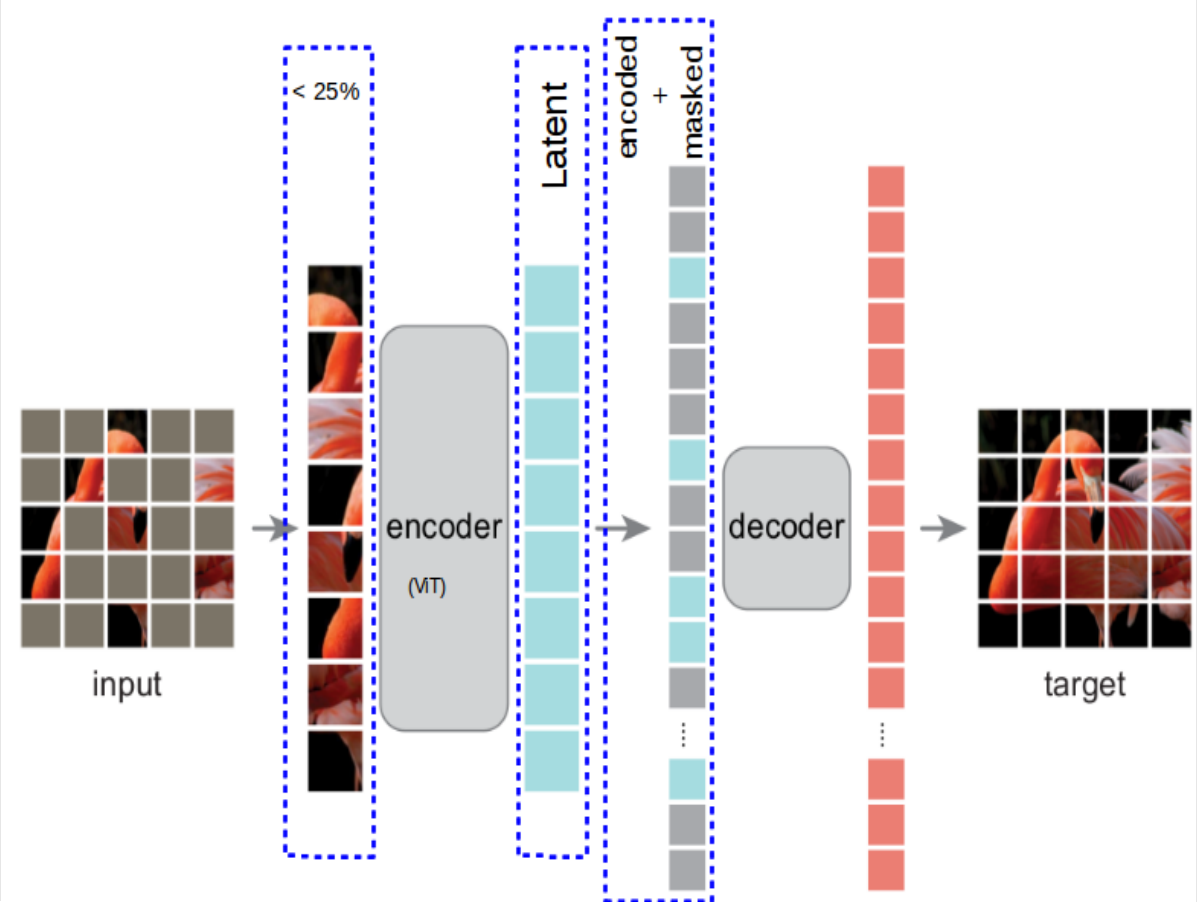

**Fig. 3 MAE architecture [7].**

### 2.3 ViTDet

ViTDet was designed to extract multi-scale features for object detection with minimal adaptation to MAE pretrained ViT. Fig. 4 shows the architecture of ViTDet building a simple feature pyramid from only the last feature map of a plain ViT backbone, and it adopts non-overlapping window attention for efficient feature extraction. To propagate information, ViTDet uses a small number of cross-window blocks, which can be implemented with global attention or convolutions. The adaptation takes place only during fine-tuning; therefore, they do not affect the upstream pretraining. The empirical study shows that ViTDet's simple design achieves good results on natural scene image object detection [5], which further proves that the general-purpose pretrained ViT from MAE can serve object detection as well. ViTDet's simple design makes it easily plug into any detector architecture. Investigating ViTDet's effectiveness for challenging aerial image object detection is the focus of this study.

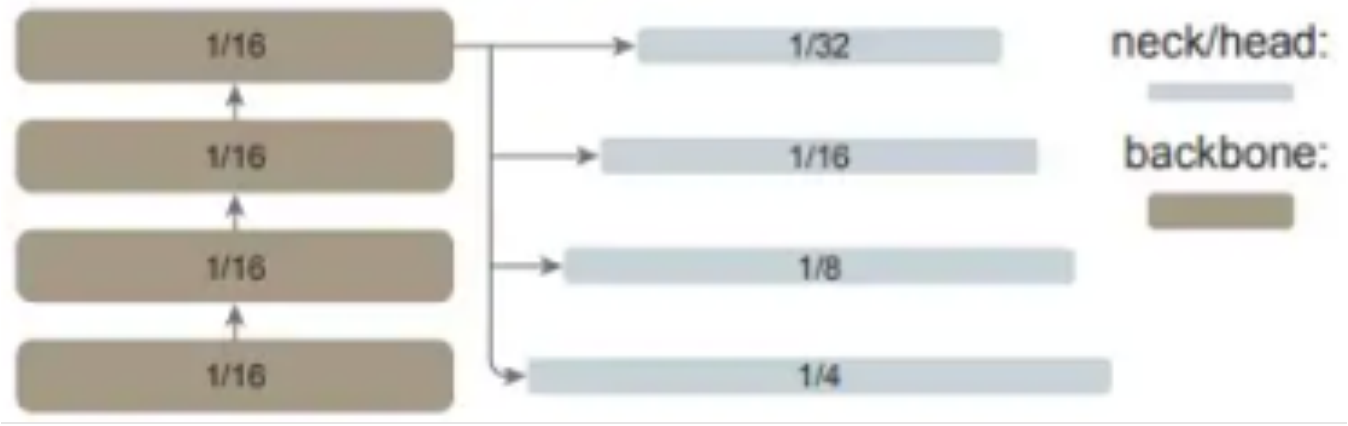

**Fig. 4 ViTDet backbone architecture. ViTDet builds a simple pyramid from only the last, large-stride (16) feature map of a plain backbone [5].**

### 2.4 Object Detection

Object detection is one of the most fundamental yet challenging CV tasks. The task is to identify and localize all the objects in an image. Each object will have a label, and its location is commonly defined by an HBB $(x, y, w, h)$, where $x$ and $y$ are center coordinates of the box, and $w$ and $h$ are width and height of the box (illustrated in Fig. 5).

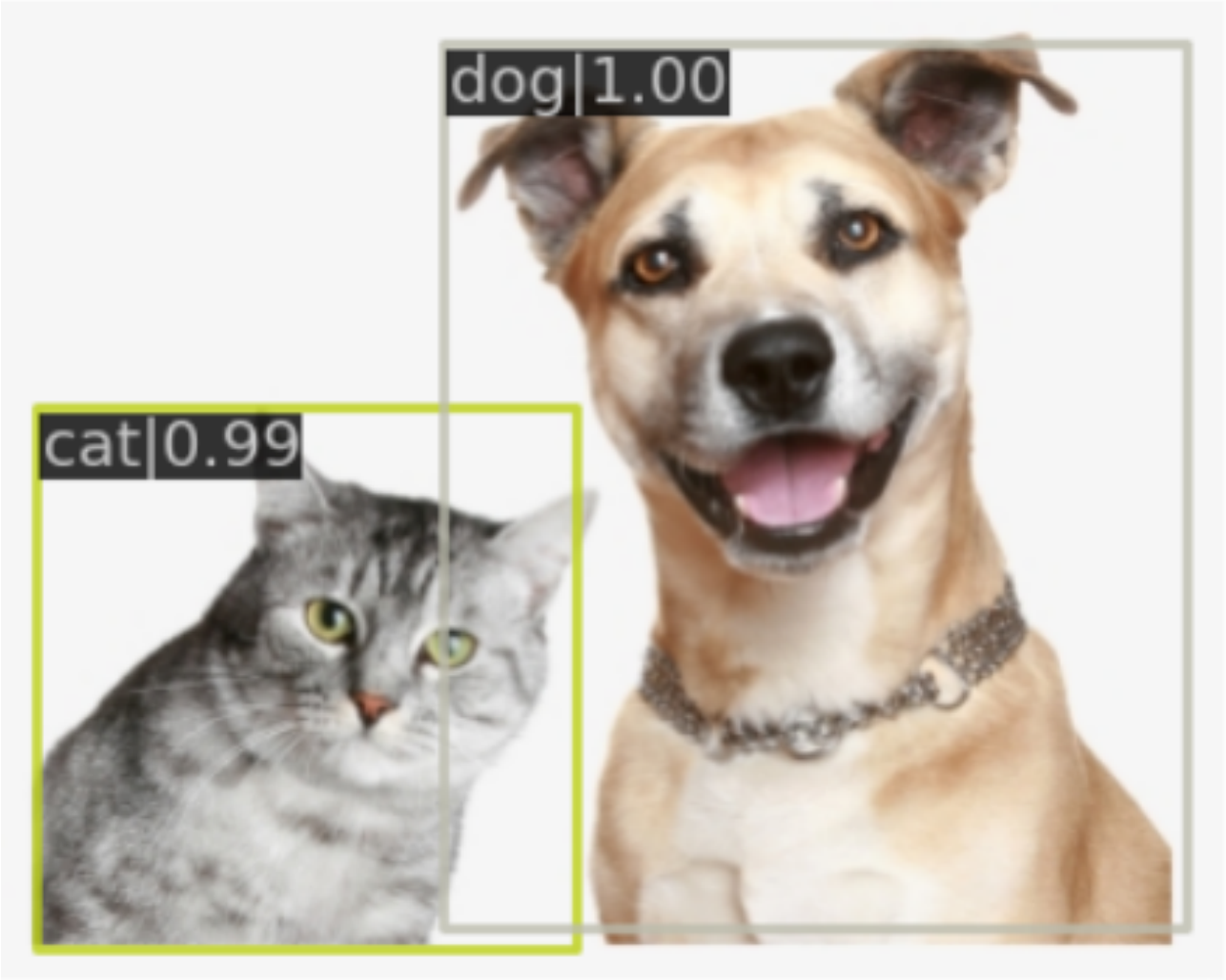

**Fig. 5 Illustration of HBB object detection.**

However, in aerial images, the objects could be arbitrarily oriented. The methods relying on HBBs often introduce mismatches between the Regions of Interest (RoI) and objects, which further deteriorate the final object classification confidence and localization accuracy [1]. For example, in Fig. 6, a RoI (top) may contain several instances, leading to difficulties for the subsequent classification and location task [1]. For this reason, research has proposed OBB annotations $(x, y, w, h, \theta)$ (see Fig. 7 for illustration), where $x$ and $y$ are center coordinates of the box and $w$, $h$, and $\theta$ are the width, height, and angle of an OBB. It should be noted that $w$ and $h$ of the OBBs are measured in different rotating coordinate systems for each object. OBBs make more accurate orientation information, especially when detecting aerial objects with a large aspect ratio, arbitrary orientation, and dense distribution. Furthermore, OBBs can also have more accurate RoIs and allow for better discriminative feature extraction for object detection. Deep learning algorithms such as oriented Region-based Convolutional Neural Network (RCNN) [160], RoI Transformer [1], and Rotation-equivalent Detector (ReDet) [161] have been proposed particularly for OBB detection. They usually adopt numerous rotated anchors with different angles, scales, and aspect ratios for better regression, resulting in significant computation burden.

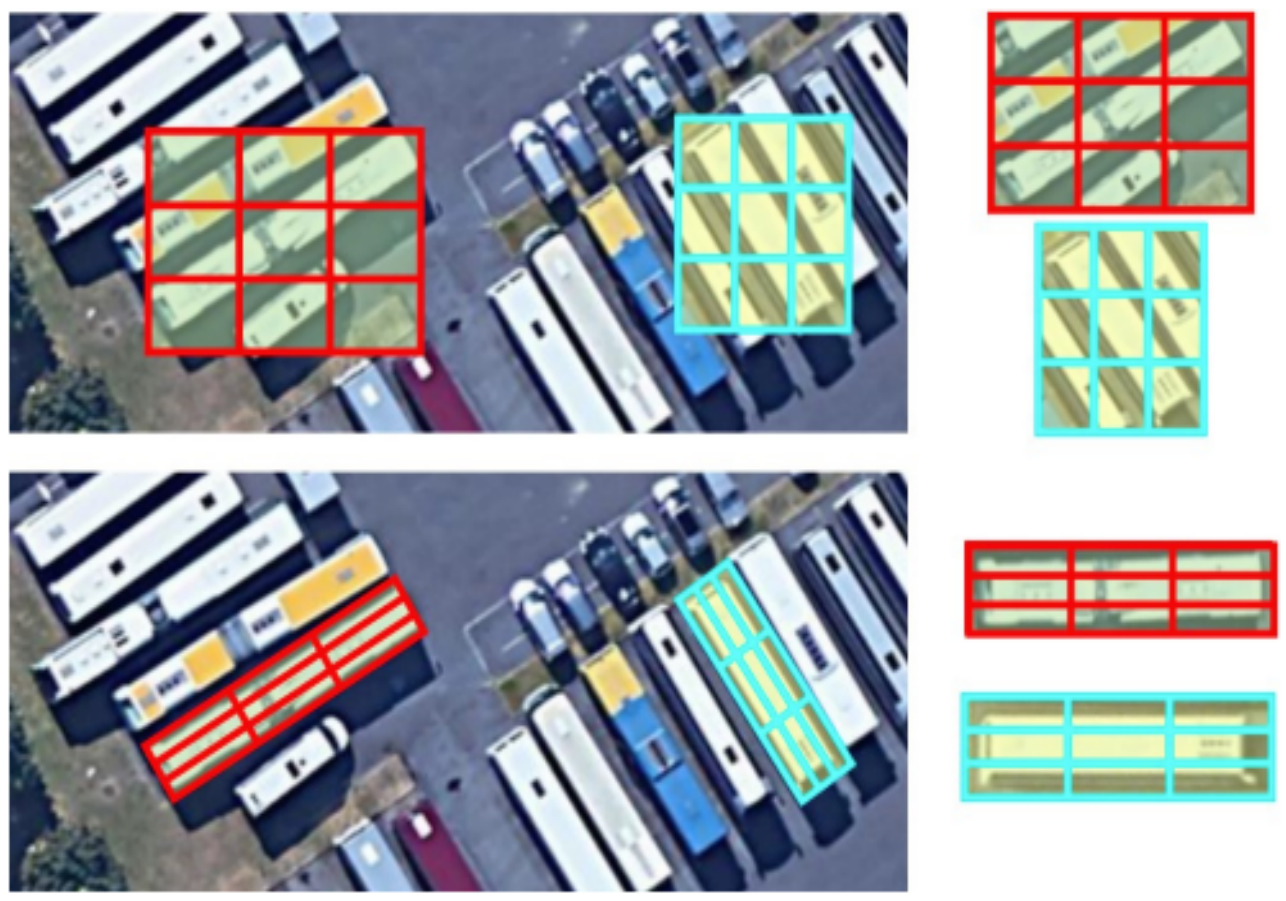

**Fig. 6 HBB (top) vs OBB (bottom) illustration in an image with many densely packed objects [1].**

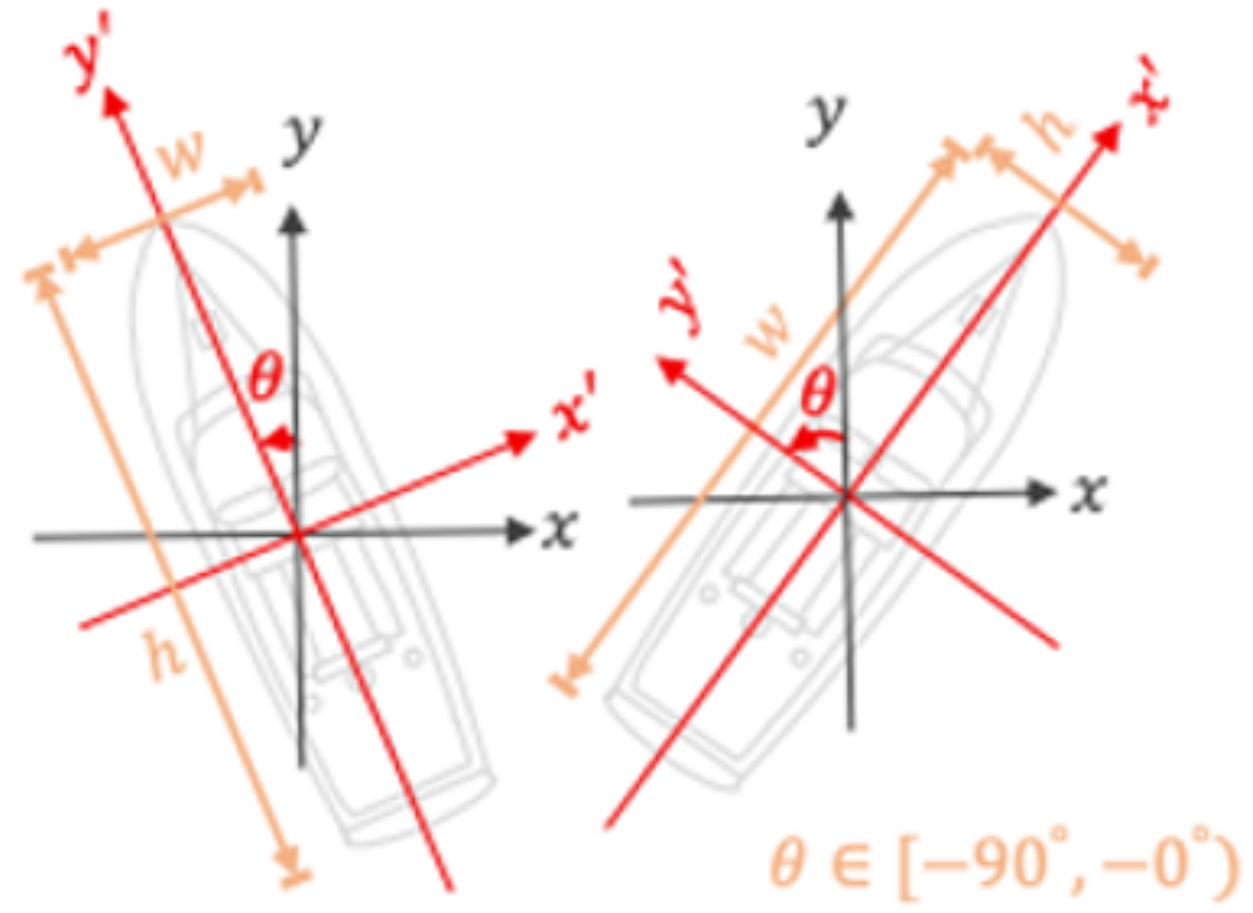

$(x, y. w, h, \theta)$

**Fig. 7 OBB definition, where x and y are center coordinates of the box and w, h, and θ are the width, height, and angle of an OBB [162].**

**2.5 Object Detection Algorithms**

As mentioned in Fig. 2, there are several types of object detection methods. Famous detection challenges have shown that two-stage methods achieve better performance than one-stage methods if heavy computation workload is not a concern. In our work, we focus on two-stage methods for their better performance. The two-stage object detection methods usually consist of the three steps proposed in RCNN [162] (illustrated in Fig. 8). The first step is region proposal, which generates a series of candidate region proposals (about 2,000) that may contain objects. The second step is feature extraction for the proposed regions. Following that, the third step is classification, where the candidate regions are distinguished as object classes or background and furtherly fine-tuned for the coordinates of the bounding boxes. As research advances, various types of algorithms have been proposed to hone the components for better performances. Next, the most fundamental deep learning method, Faster RCNN [163], and the algorithms explored in this research will be presented.

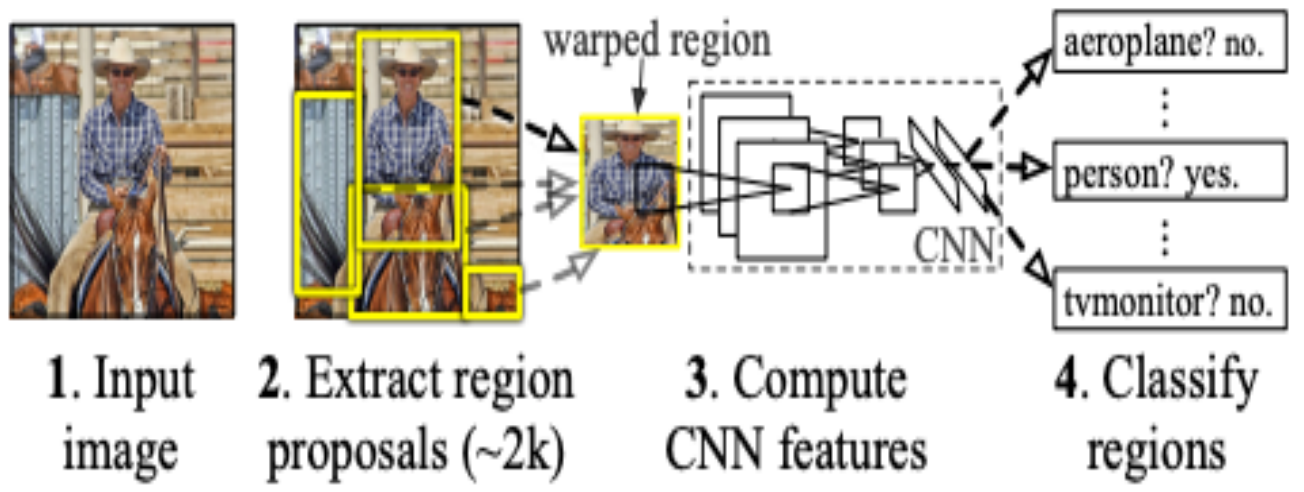

**Fig. 8 Object detection system overview [163].**

**2.6 Faster RCNN**

Faster RCNN [163] is the first end-to-end object detection method fully using deep learning techniques, which is more efficient than its predecessors, RCNN [162] and Fast RCNN [164]. Faster RCNN proposes a novel idea called region proposal network (RPN), which fully utilizes convolutional layers extracted features to generate proposals. Compared with conventional region proposal generation algorithms like Selective Search [165], which is an offline algorithm and makes it impossible to train whole algorithm from end to end, RPN is much more efficient. After RPN, Faster RCNN then uses the RoI pooling layer to extract a fixed-length feature vector from each region proposal. Fig. 9 depicts the architecture of Faster RCNN and the sequential relationship among backbone (convolutional layers), RPN, and the RoI pooling layer. Based on Faster RCNN, several variants have been proposed to improve the performance of object detection. Next, we present the relevant methods tested in our study.

**2.7 Mask RCNN**

Mask RCNN [166] is an extension of Faster RCNN. Besides Faster RCNN's two outputs for each candidate object—a class label and a bounding box—a third type of output, object mask, is proposed. Fig. 10 illustrates the architecture of Mask RCNN. The backbone of Mask RCNN is for feature extraction, and it can be traditional ResNet [2], Swin-T [4], or newly proposed ViTDet [5]. RPN is the same as the one in Faster RCNN. The novel element of Mask RCNN is the RoIAlign layer, which can preserve the pixel-level spatial correspondence and address the shortfalls of Fast/Faster RCNN. The mask head is a Fully Convolutional Network (FCN) [167] on top of a feature map. Mask RCNN is still simple to train and generalizes well, but it introduces a small computation overhead to Faster RCNN.

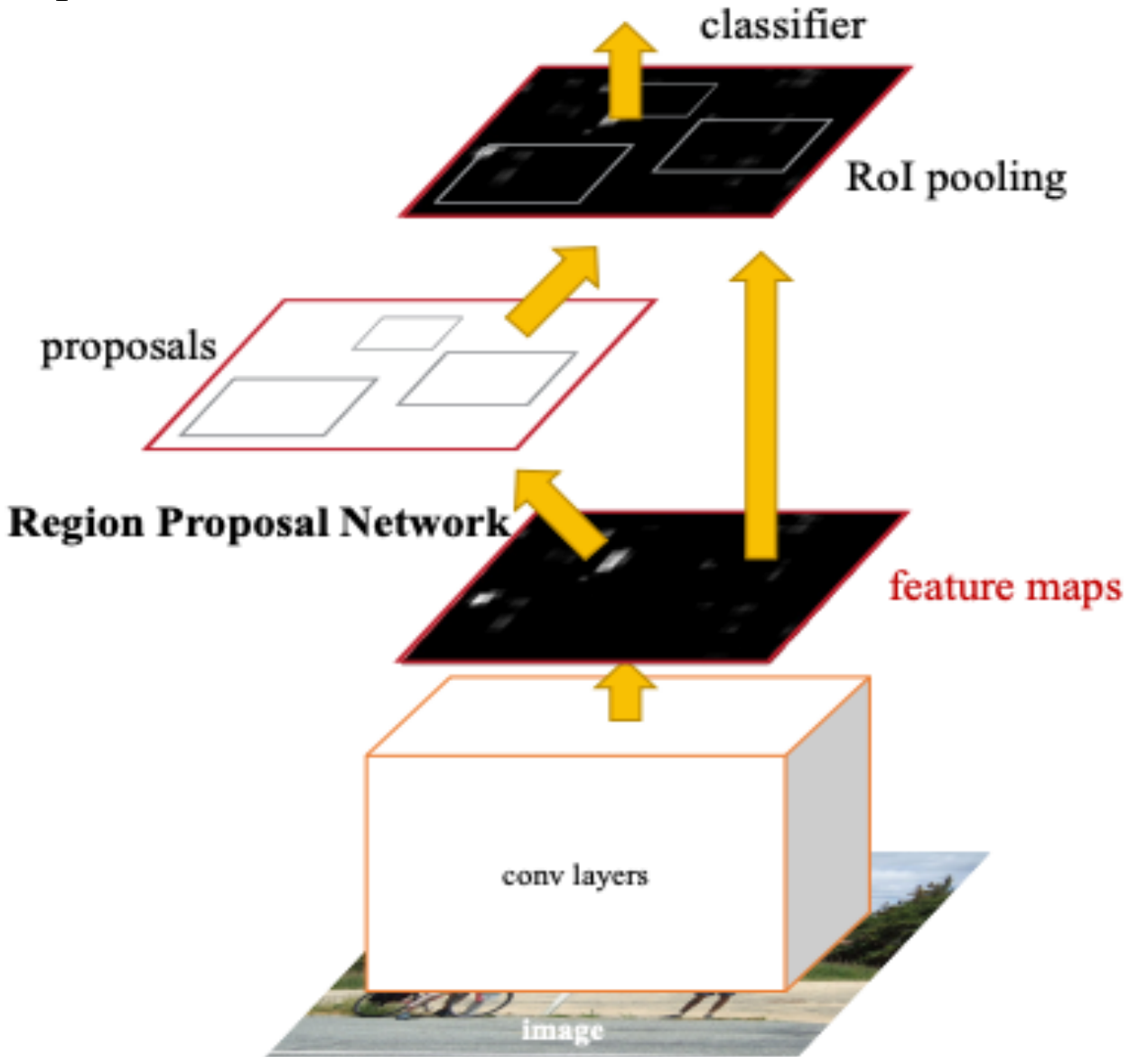

**Fig. 9 Faster RCNN architecture [164].**

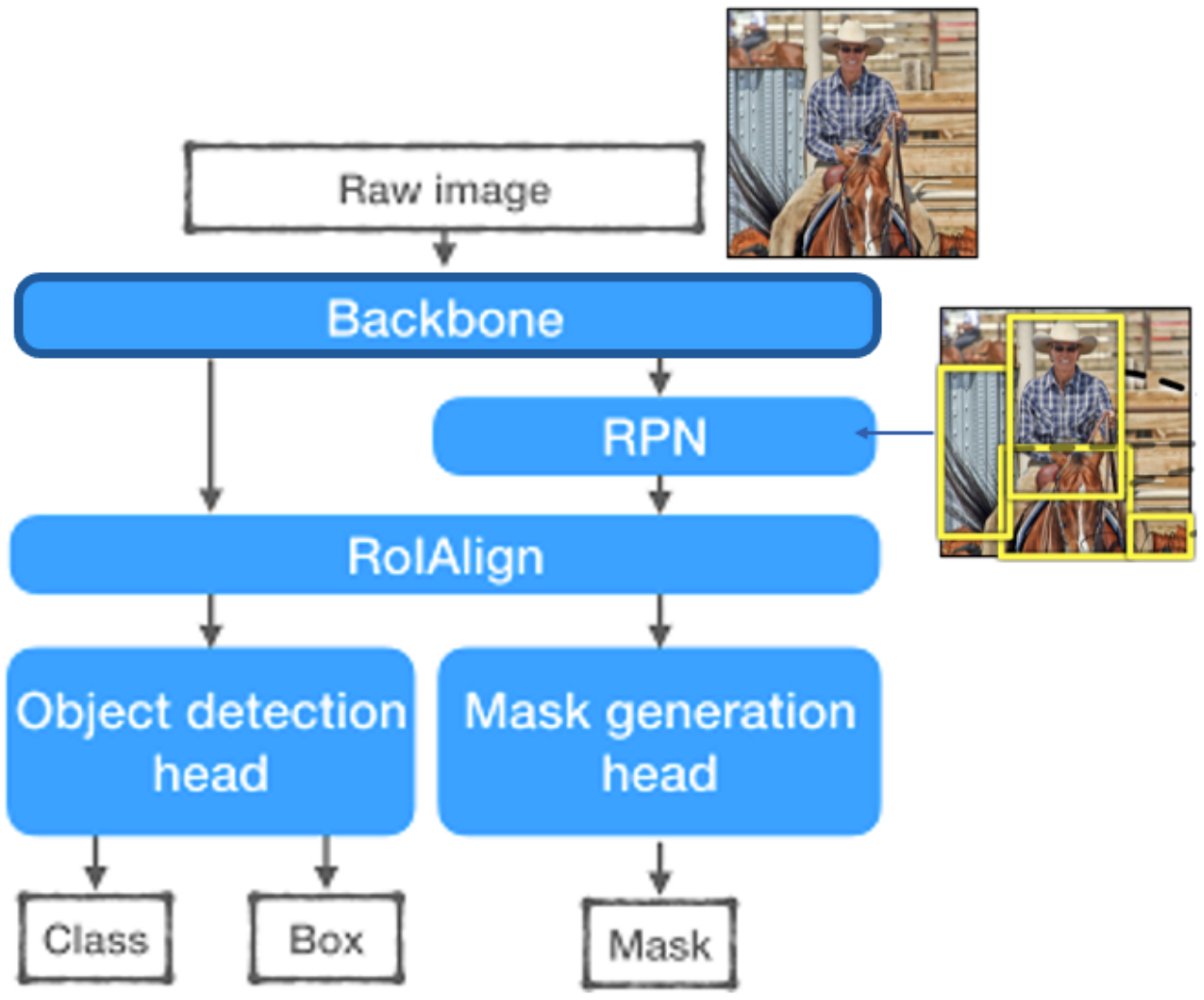

**Fig. 10 Mask RCNN architecture [169].**

### 2.8 Cascade RCNN

Cascade RCNN [168] adopts a new trick for better performance, classifying with multistage classifiers. The trick works in such a way that early stages can discard many easy negative samples; therefore, later stages can focus on handling more difficult examples. Fig. 11 illustrates the architecture of Cascade RCNN, where "I" is input image, "conv" is the convolutions backbone, "pool" is for the region-wise feature extraction, "H" represents various network head, "B" is the bounding box, "C" is classification, and "B0" is proposals in all architectures. An object detection architecture like Faster RCNN can be deemed as a cascade (i.e., the RPN removing large amounts of background and the following detector head classifying the remaining regions). Therefore, Cascade RCNN extends the idea to multiple stages in the classification layer to enhance the performance. When mask head is also included in the output, the algorithm is called Cascade Mask RCNN, which is used for HBB object detection in our study.

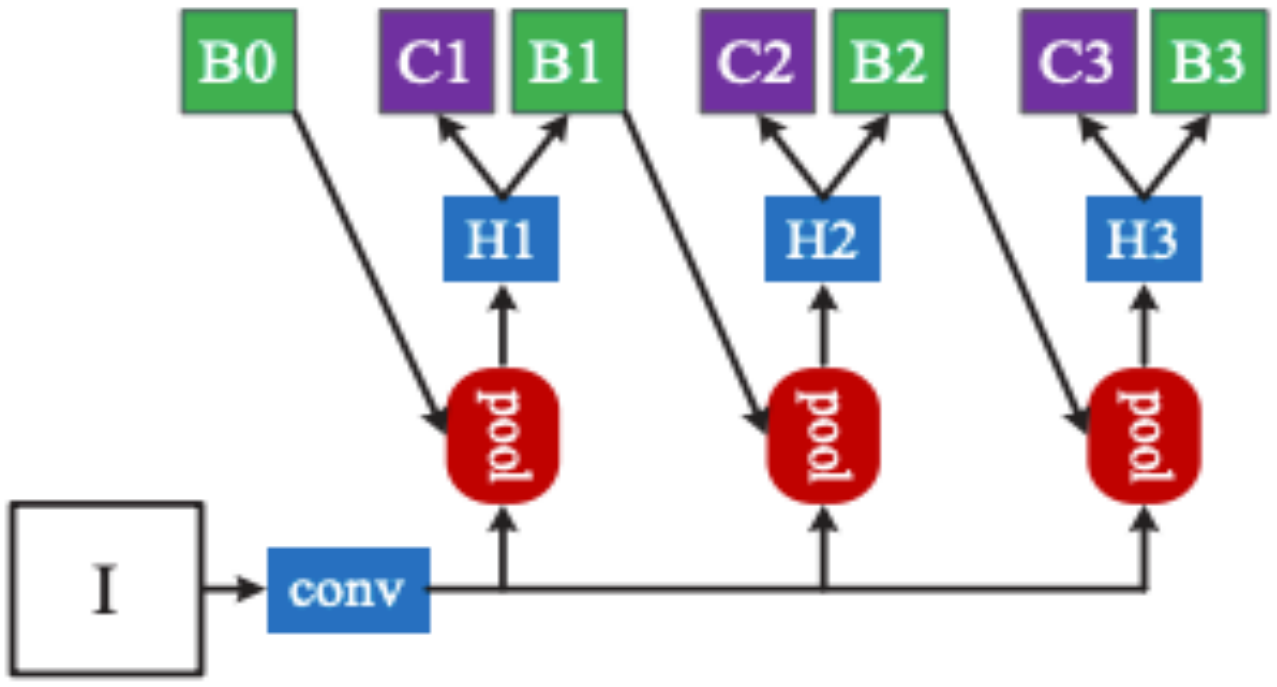

**Fig. 11 Cascade RCNN architecture [170].**

### 2.9 RoI Transformer

RoI Transformer [1] was designed specifically for OBB object detection. In the past, rotated anchors have been used to tackle the OBB object detection problem. The design always multiplies the number of anchors, which considerably increases the computation burden. Hence, RoI Transformer was tried for reducing the computation burden. Fig. 12 illustrates the architecture of RoI Transformer. In specific, it first adopts a Rotated Region of Interest (RRoI) learner to transform a Horizontal Region of Interest (HRoI). Based on the RRoIs, it then uses a Rotated Position Sensitive RoI Align (RPS-RoI-Align) module to extract rotation-invariant features, which are then used for enhancing subsequent classification and regression performance. RoI Transformer is a light-weighted component and can be easily plugged into any detector framework for OBB object detection.

### 2.10 Rotation-Equivalent Detector (ReDet)

ReDet [161] was also proposed to solve OBB aerial image object detection problems. It introduces rotation-equivariant networks into the detector to extract rotation-equivariant features, which can accurately predict the orientation and result in a huge reduction in model size. Fig. 13 illustrates the working mechanism of ReDet. Fig. 13a shows the overall architecture of ReDet, which first uses the rotation-equivariant backbone to extract rotation-equivariant features, followed by an RPN and RoI Transformer (RT) [1] to generate RRoIs. After that, a novel Rotation-Invariant RoI Align (RiRoI Align) is used to produce rotation-invariant features for RoI-wise classification and bounding box regression. Fig. 13b shows rotation-equivariant feature maps. Under the cyclic group $C_N$, the rotation-equivariant feature maps with the size $(K, N, H, W)$ have $N$ orientation channels, and each orientation channel corresponds to an element in $C_N$. Fig. 13c illustrates RiRoI Align. The proposed RiRoI Align consists of two parts: spatial alignment and orientation alignment. For an RRoI, spatial alignment warps the RRoI from the spatial dimension, while orientation alignment circularly switches orientation channels and interpolates features to produce completely rotation-invariant features. ReDet has achieved state-of-the-art performance [169]; therefore, it was selected in our study to test ViTDet for OBB object detection. Next, we give more details about aerial image datasets to test ViTDet backbone.

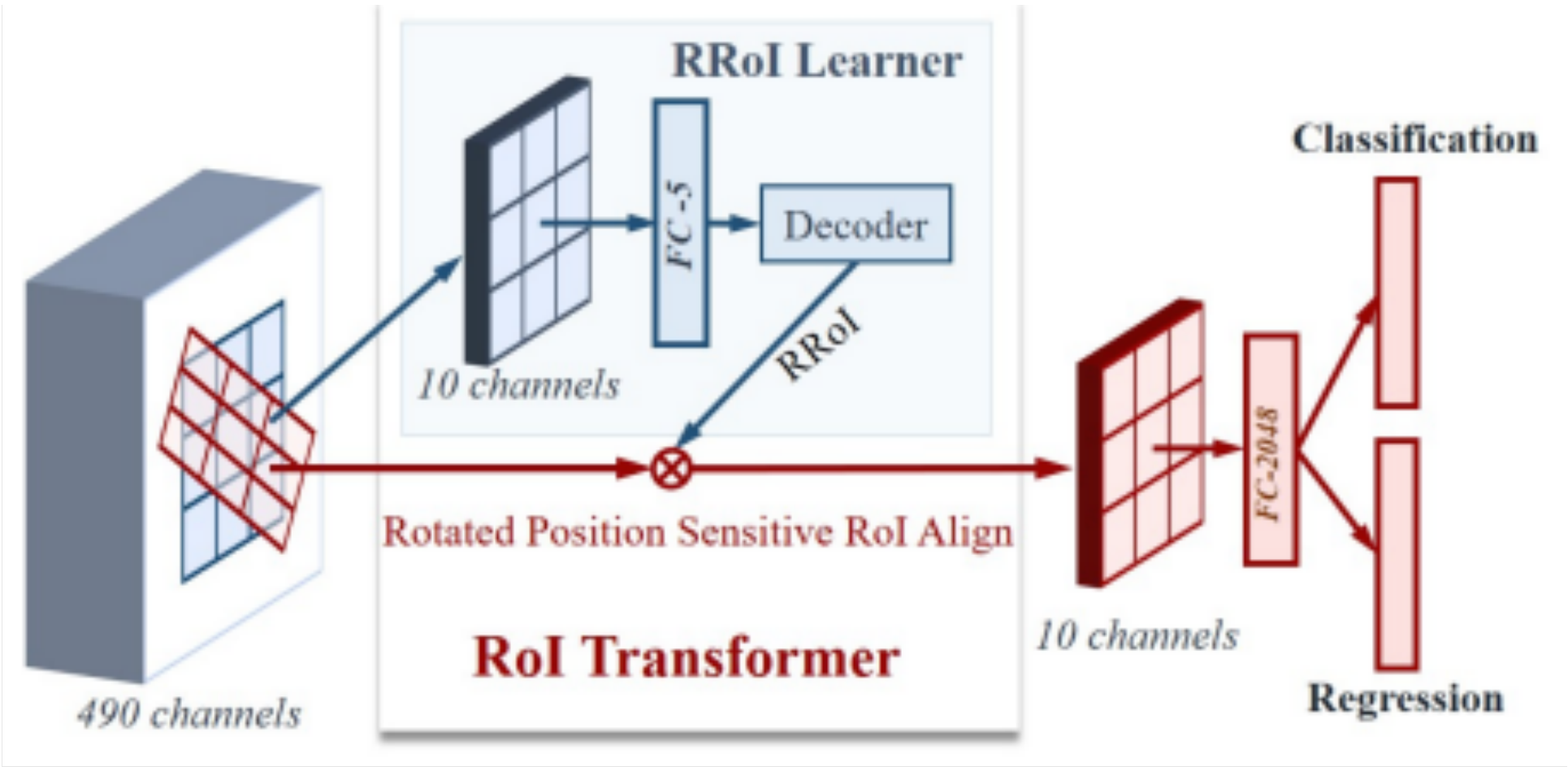

**Fig. 12 Architecture of RoI Transformer [1].**

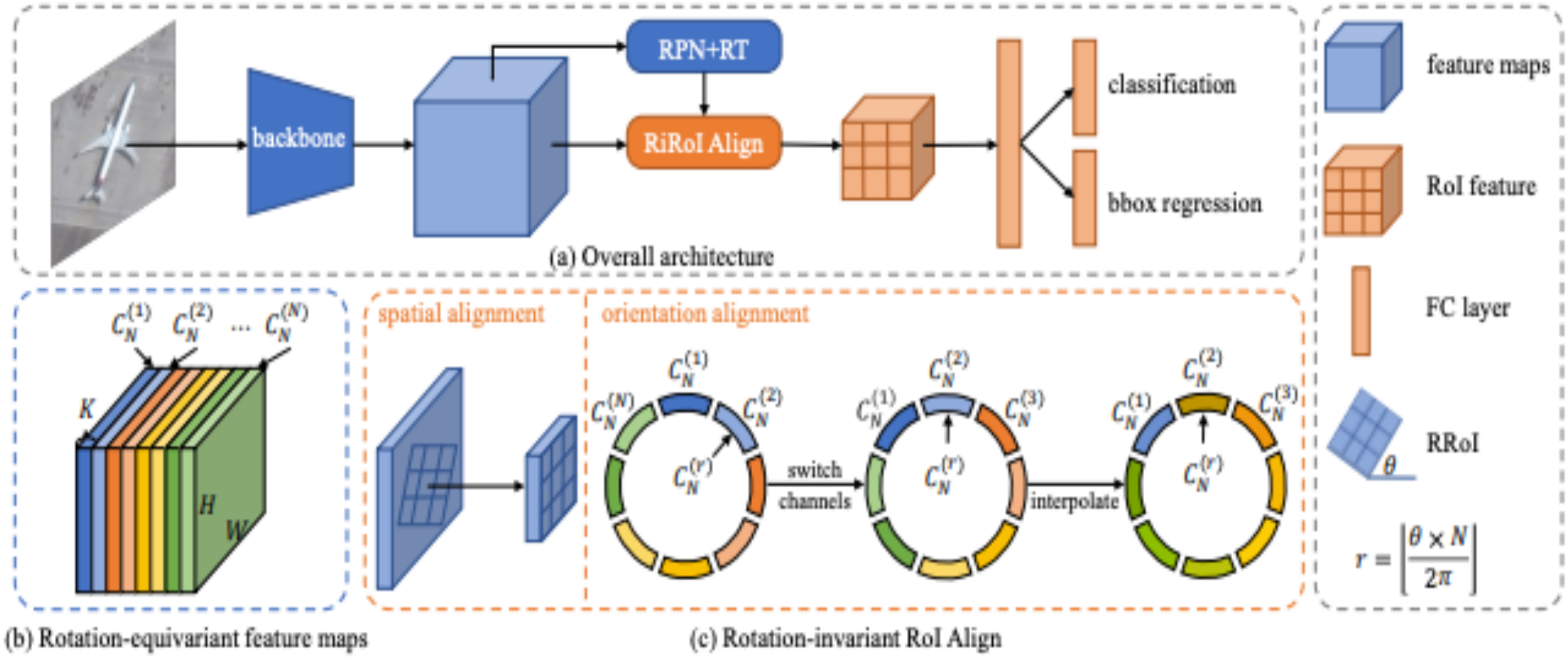

**Fig. 13 ReDet architecture [161].**

## 3. DATASETS

To evaluate the newly proposed backbone of MAE pretrained ViTDet for aerial image object detection, we conducted 25 experiments across three distinct datasets of aerial images: 1) Airbus Aircraft Detection [170], 2) RarePlanes [171], and 3) Dataset of Object DeTection in Aerial images (DOTA) [172]. The smallest dataset is Airbus Aircraft Detection, with 103 images, and the largest dataset is RarePlanes, with about 68,000 images. These two both use HBB annotations. DOTA is the most complicated dataset with OBB annotations. Table 2 gives the details of the three datasets. A short introduction about each dataset will be provided next.

### 3.1 Airbus Aircraft Detection

The Airbus Aircraft Detection [173] dataset is collected from Airbus' Pleiades twin satellites for earth observation, which collect pictures of airports worldwide on a regular basis. This dataset contains 103 images with 0.5 m resolution (see Fig. 14 for an example). Each image is stored as a JPEG file of size 2,560 x 2,560 pixels (i.e., 1,280 meters x 1,280 meters). Some airports could appear multiple times at different acquisition dates. Some images may include fog or cloud because of weather. The annotations are provided in the form of closed GeoJSON polygons. A CSV file named annotations.csv provides all annotations—one annotation per line with the corresponding object ID; filename as image ID; annotation box; and class label, mainly Aircraft (3,316 instances) or Truncated_Aircraft (109 instances) when an aircraft is located at the border of an image. The minimum and maximum number of aircraft in an image are 5 and 92, respectively.

### 3.2 RarePlanes

RarePlanes [171] is an open-source dataset that includes both real and synthetically generated satellite images. The RarePlanes dataset is specifically designed to automatically detect aircraft and their attributes in satellite images (see Fig. 15 for examples). To date, RarePlanes is the largest openly available high-resolution dataset created to test the value of synthetic data from an overhead perspective. The real images are collected from 253 Maxar WorldView-3 satellite scenes, spanning 112 locations and 2,142 $km^2$ with 14,700 hand-annotated aircraft. The accompanying synthetic dataset is generated via AI.Reverie's simulation platform and has about 60,000 synthetic satellite images covering a total area of 9,331 $km^2$ with about 630,000 aircraft annotations.

Both the real and synthetically generated aircraft have been given 10 fine-grained attributes—aircraft length, wingspan, wing shape, wing position, wingspan class, propulsion, number of engines, number of vertical stabilizers, presence of canards, and aircraft role—which are derived from the previous nine attributes. Seven role classes have been defined; Table 3 summarizes aircraft role count for real dataset, in which the first column lists seven "aircraft role" classes. As demonstrated in Table 3, the most common aircraft role is Small Civil Transport/Utility, and the least common one is Military Bomber. More detail on role definitions can be found in the "RarePlanes User Guide" at https://www.cosmiqworks.org/rareplanes-public-user-guide/. We conducted two types of object detection tasks—aircraft and aircraft role—on both sub-datasets to evaluate MAE pretrained ViTDet backbone's performance.

### 3.3 DOTA

DOTA [174] is the largest aerial image dataset for OBB object detection (see Fig. 16 for some examples), and it is deemed as the most challenging dataset in the earth observation community for its various image sizes and densely packed objects. It has released three different versions. DOTA-v1.0 contains 2,806 aerial images, with the size ranging from 800 × 800 to 4,000 × 4,000 and containing 188,282 instances. DOTA-v1.0 has 15 common categories: Plane (PL), Baseball diamond (BD), Bridge (BR), Ground track field (GTF), Small vehicle (SV), Large vehicle (LV), Ship (SH), Tennis court (TC), Basketball court (BC), Storage tank (ST), Soccer-ball field (SBF), Roundabout (RA), Harbor (HA), Swimming pool (SP), and Helicopter (HC). The second version DOTA-v1.5 was released for 2019 Detecting Objects in Aerial Images (DOAI) Challenge. Compared with v1.0, it has an extra category, Container crane, and more extremely small instances (less than 10 pixels), resulting in 402,089 instances. The third version, DOTA-v2.0, collects more aerial images from Google Earth and GF-2 Satellite. DOTA-v2.0 has 18 categories, 11,268 images, and 1,793,658 instances. Compared with DOTA-v1.5, it further adds the new categories of Airport and Helipad. Our study focused on DOTA-v1.0 due to abundant baseline benchmarks available for evaluating ViTDet's performance.

**Table 2 Tested aerial image datasets**

| Datasets | Subsets | Tasks | # Object Types | # Images | # Instances | Image Width | Annotation | Year Available |
|---|---|---|---|---|---|---|---|---|
| Airbus Aircraft Detection | - | Aircraft | 2 | 103 | 3,425 | 2,560 | HBB | 2021 |
| RarePlanes | Real | Aircraft | 1 | 8,527 | 14,700 | 512 | HBB | 2020 |
| | Synthetic | Aircraft | 1 | 60,000 | 629,551 | 1,920 | | |
| | Real | Aircraft role | 7 | 8,527 | 14,700 | 512 | | |
| | Synthetic | Aircraft role | 7 | 60,000 | 629,551 | 1,920 | | |
| DOTA | v 1.0 | Objects | 15 | 2,806 | 188,282 | 800-4,000 | OBB | 2018 |

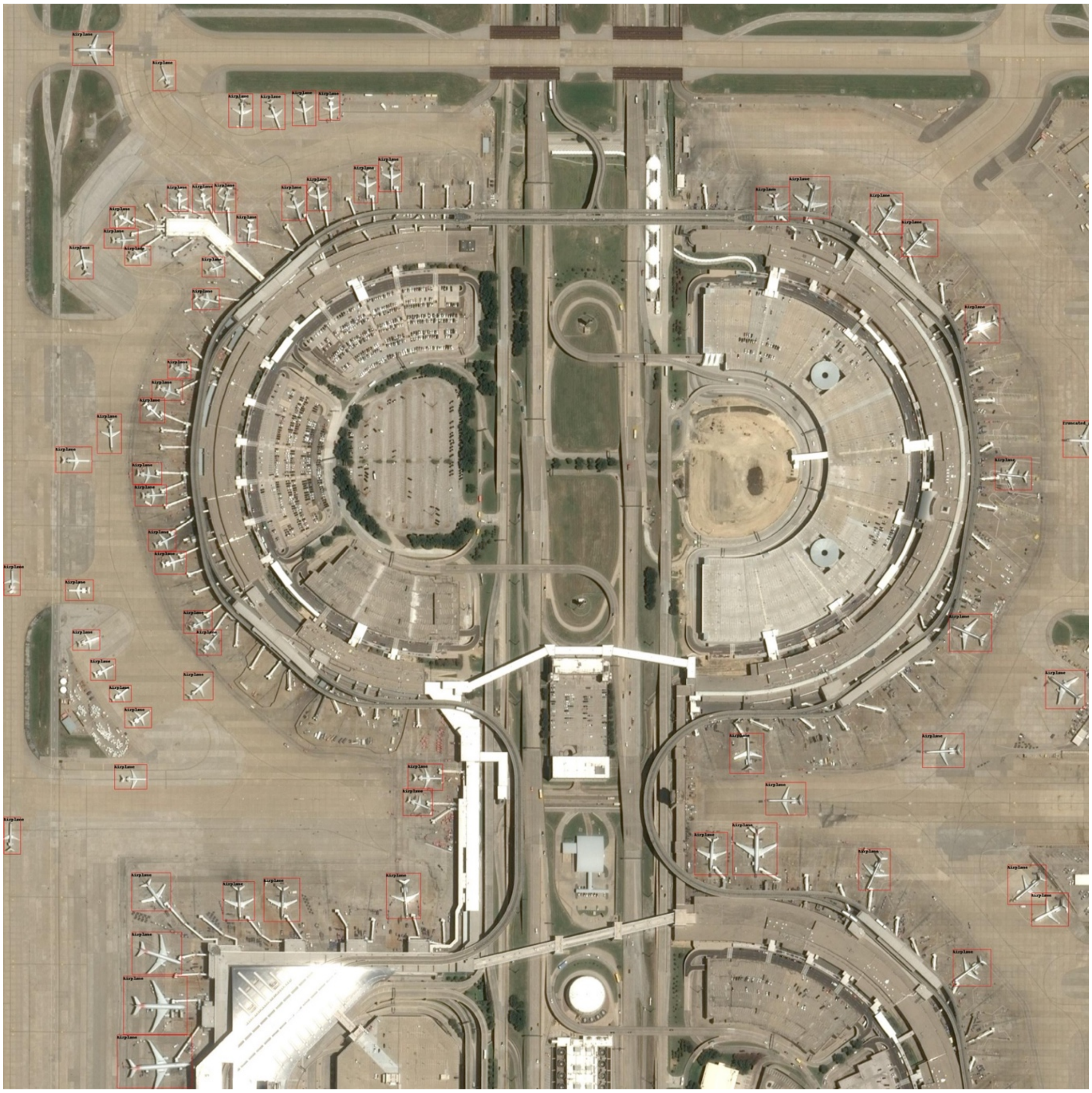

**Fig. 14 Airbus Aircraft Detection image example.**

**Table 3  Real dataset role count**

| Aircraft role | Count |
|---|---|
| Small Civil Transport/Utility | 8002 |
| Medium Civil Transport/Utility | 5132 |
| Large Civil Transport/Utility | 1098 |
| Military Transport/Utility/AWAC | 283 |
| Military Fighter/Interceptor/Attack | 171 |
| Military Trainer | 15 |
| Military Bomber | 6 |

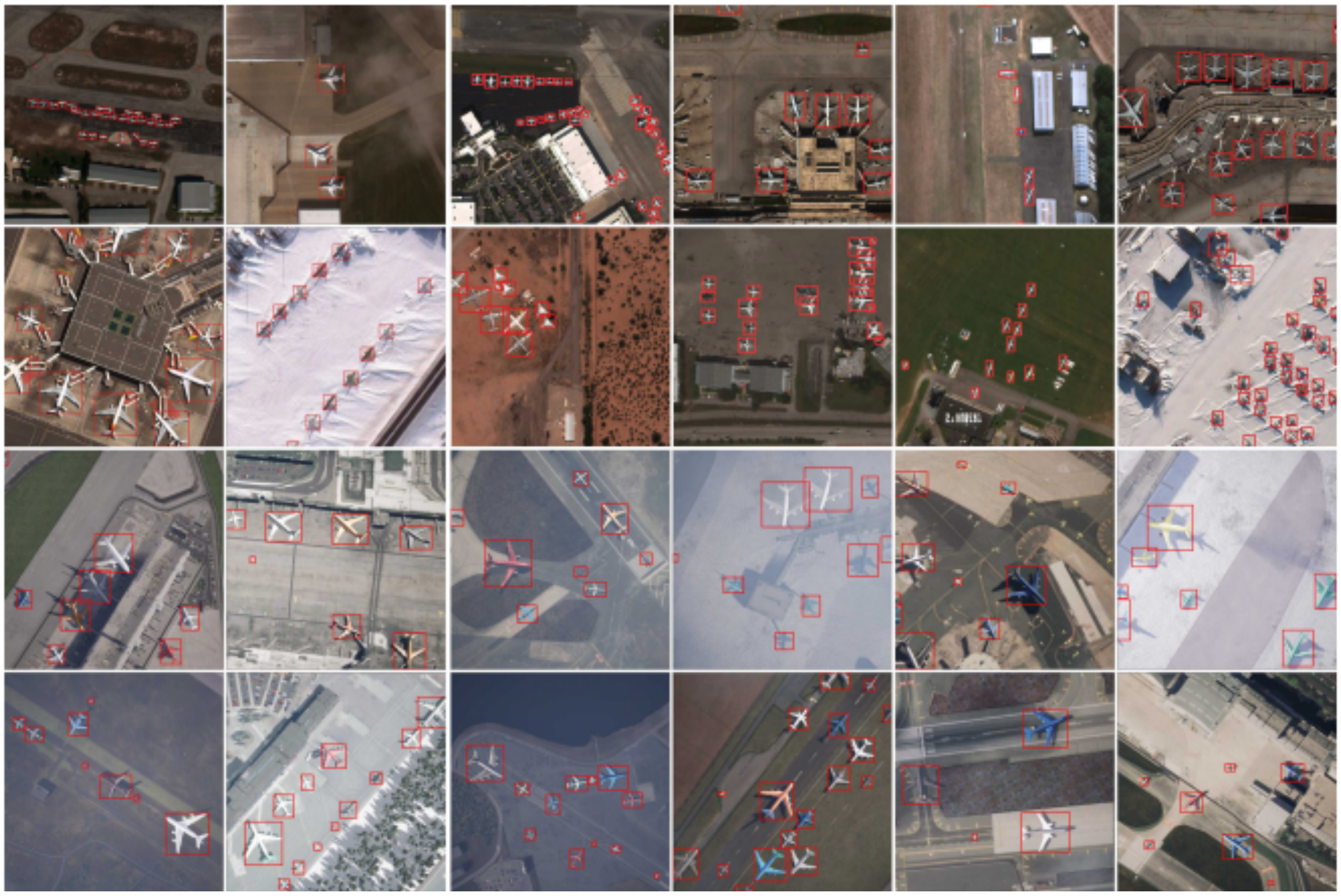

Fig. 15 Examples of the real and synthetic datasets present in RarePlanes [173].

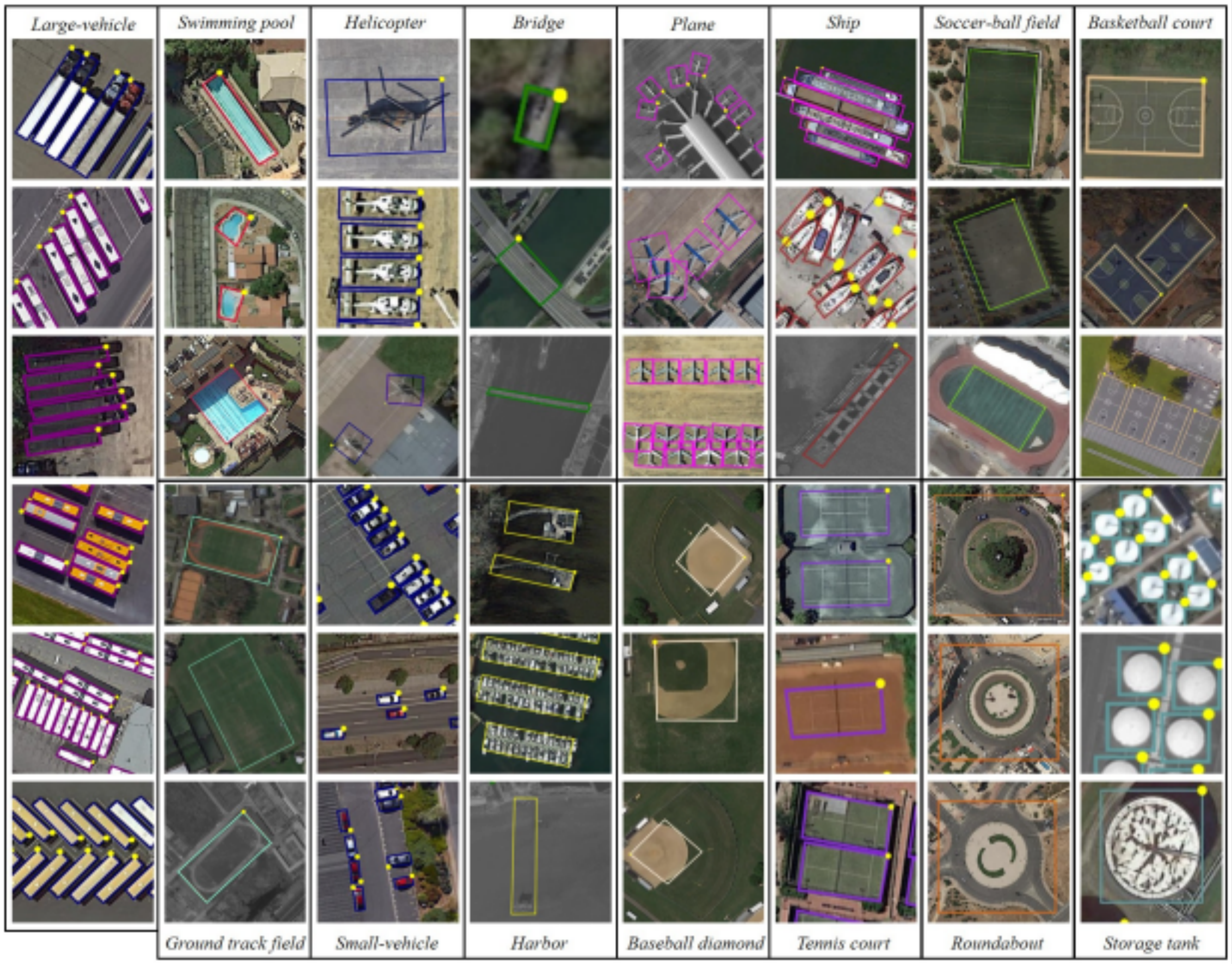

Fig. 16 Examples of annotated images in DOTA [176].

## 4. IMPLEMENTATION PLATFORMS

To evaluate the new backbone ViTDet in the aforementioned algorithms, we chose two well-known platforms in the CV field: Detectron2 [175] and MMRotate [169]. Detectron2 is the official implementation site for ViTDet and is used for HBB object detection. MMRotate is selected because it has most of state-of-the-art algorithms for OBB object detection, which Detectron2 lacks.

### 4.1 Detectron2

Detectron2 [175] is an open-source research platform developed by Facebook AI Research [175]. The platform is implemented in PyTorch. It provides many state-of-the-art detection and segmentation algorithms, including FPNs, numerous variants of the pioneering Mask RCNN model family, and the latest MAE pretrained ViTDet backbone. Therefore, we used Detectron2 to implement aerial image HBB object detection with its provided pretrained models.

### 4.2 MMRotate

OpenMMLab [176] is another open-source platform to provide powerful CV packages like Detectron2. For general HBB object detection, MMDetection in OpenMMLab is the go-to package and forms the basis for MMRotate [169], which is specially designed for OBB object detection. According to Table 4 provided by Zhou et al. [169], MMRotate provides 18 OBB algorithms and four famous datasets. In addition, its modular design with multiple choices of orientation angle representations, backbones, necks, and detection heads makes it very easy and flexible to set up a new model. For example, it can support multiple angle representations. Popular OpenCV definition, long edge 90° definition, and long edge 135° are all supported in MMRotate. MMRotate also provides baseline benchmarks for comparison. Therefore, we selected MMRotate for customization of RoI Transformer and ReDet, where ViTDet will be used as the backbone. Note that at the time of this research, ViTDet has not officially been implemented in MMRotate. We used a non-official version of ViTDet from [177] for OBB object detection.

## 5. RESULTS

This section presents the experiment results of aerial image object detection using the MAE pretrained ViTDet backbone.

### 5.1 Experimental Setup

To make a comprehensive evaluation, we conducted 25 experiments on the selected three datasets: 1) Airbus Aircraft Detection [170], 2) RarePlanes [171], and 3) DOTA [172]. The experiments tested three types of backbones (i.e., ResNet [2], Swin Transformer [4], and ViTDet [5]) in four object detection algorithms (i.e., Mask RCNN [166], Cascade Mask RCNN [168], RoI Transformer [1], and ReDet [161]). For the Airbus Aircraft and RarePlanes datasets, we tested Mask RCNN and Cascade Mask RCNN algorithms on the Detectron2 [175] platform. For the DOTA dataset, we tested RoI Transformer and ReDet on the MMRotate [169] platform. The MAE pretrained ViTs were downloaded from https://github.com/facebookresearch/mae [178]. All the experiments were carried out on four A100 GPUs with 250 GB memory. More specific implementation details for each dataset will be presented in the corresponding sections.

### 5.2 Evaluation Metrics

Average precision (AP) is a commonly used metric to evaluate object detection algorithms, and it is derived from precision and recall. There are several variants of AP. Different platforms may adopt different versions of AP. In details, Detectron2 uses COCO-defined AP metrics (see Table 5 for the detailed list), which mainly focus on the accuracy of the bounding box. COCO-defined AP is averaged across all classes and 10 Intersection Over Union (IOU) values ranging from 0.5 to 0.95 in steps of 0.05 [155]. By contrast, in the MMRotate platform, AP is calculated separately for each class, and mean AP (mAP) is calculated by averaging AP over all classes. To have a fair comparison, we calculated the default evaluation metrics defined by the two platforms.

**Table 4 Open source rotated object detection benchmarks [169]**

| Benchmark | AerialDet | JDet | OBBDet | AlphaRotate | MMRotate |
|---|---|---|---|---|---|
| DL library | PyTorch | Jittor | PyTorch | TensorFlow | PyTorch |
| Inference engine | PyTorch | Jittor | PyTorch | TensorFlow | PyTorch onnx runtime |
| OS | Linux | Windows Linux | Windows Linux | Linux | Windows Linux |
| Algorithm | 5 | 8 | 9 | 16 | 18 |
| Dataset | 1 | 4 | 5 | 11 | 4 |
| Doc. | - | - | - | ✓ | ✓ |
| Easy install | - | - | - | - | ✓ |
| Maintain | - | ✓ | ✓ | ✓ | ✓ |

**Table 5 COCO-defined AP evaluation metrics [154], used in default by Detectron2**

```
Average Precision (AP):
  AP              % AP at IoU=.50:.05:.95 (primary challenge metric)
  AP^IoU=.50      % AP at IoU=.50 (PASCAL VOC metric)
  AP^IoU=.75      % AP at IoU=.75 (strict metric)
AP Across Scales:
  AP^small        % AP for small objects: area < 32²
  AP^medium       % AP for medium objects: 32² < area < 96²
  AP^large        % AP for large objects: area > 96²
```

### 5.3 Airbus Aircraft Object Detection Results

To detect aircraft in this small dataset, we have taken the following three steps:

**Step 1. Data preparation**

- Split dataset (103 images) into training (92 images) and testing (11 images) subsets.
- Convert he data into COCO format for easy use of ViTDet in Detectron2 packages.

**Step 2. Experiment setup**

- Downloaded COCO pretrained models of Mask RCNN and Cascade Mask RCNN from the website https://github.com/facebookresearch/detectron2/tree/main/projects/ViTDet.
- Set up the configuration files for model training.

**Step 3. Model fine-tuning**

Table 6 shows the experiments conducted and the performance evaluation results. The tested backbones are as follows: ResNeXt-101 [156] is a convolutional neural network (CNN) backbone with 101 layers and is pretrained in a supervised manner. ViTDet, ViT-L is ViTDet backbone built with a large version of ViT that has 24 layers and 1024-dimension output. ViTDet, ViT-H is ViTDet backbone built with a huge version of ViT that has 32 layers and 1280-dimension output. The column FT-epoch is the epochs for fine-tuning.

Yellow color highlights the best metrics in Table 6. As expected, Cascade Mask RCNN performs better than Mask RCNN; larger backbones achieve better performance. Cascade Mask RCNN with backbone of ViTDet, ViT-H achieves the best performance in all evaluation metrics except for AP75, an evaluation metric when IOU equals 0.75. Most importantly, ViTDet outperforms ResNeXt-101 in most of evaluation metrics, and ResNeXt-101 is deemed as one of top CNN backbones. According to Table 6, ViTDet performs much better (20-50% improvement) than ResNeXt-101 on APs, which measures AP for small object detection. For APl, a metric to measure AP for large object detection, ViTDet also beats ResNeXt-101 by a large margin of 16-20%. For AP, 6-10% improvement has been achieved by ViTDet. In short, the new backbone ViTDet greatly improves object detection performance on this small dataset.

Fig. 17 shows object detection results on a testing image. There are about 90 aircraft in this testing image; all but three are detected and one is falsely labeled. Therefore, ViTDet backbone does a good job for this testing image.

**Table 6. Airbus Aircraft object detection results comparison.** Note APs, APm, and APl represent COCO-defined $AP^{small}$, $AP^{medium}$, and $AP^{large}$ listed in Table 5, respectively.

| Method | Backbone | Pre-train | FT-epoch | Learning rate | AP | AP50 | AP75 | APs | APm | APl |
|---|---|---|---|---|---|---|---|---|---|---|
| Mask RCNN | ResNeXt-101 | IN1K, sup | 1000 | 0.00010 | 48.36 | 72.91 | 64.21 | 0.00 | 47.76 | 52.56 |
| Mask RCNN | ViTDet, ViT-L | IN1K, MAE | 1000 | 0.00010 | 54.80 | 79.55 | 62.93 | 20.00 | 50.56 | 69.49 |
| Cascade Mask RCNN | ViTDet, ViT-L | IN1K, MAE | 1000 | 0.00010 | 57.08 | 80.38 | 75.63 | 50.00 | 53.11 | 68.64 |
| Cascade Mask RCNN | ViTDet, ViT-H | IN1K, MAE | 750 | 0.00001 | 59.75 | 83.62 | 67.50 | 50.00 | 55.25 | 73.21 |

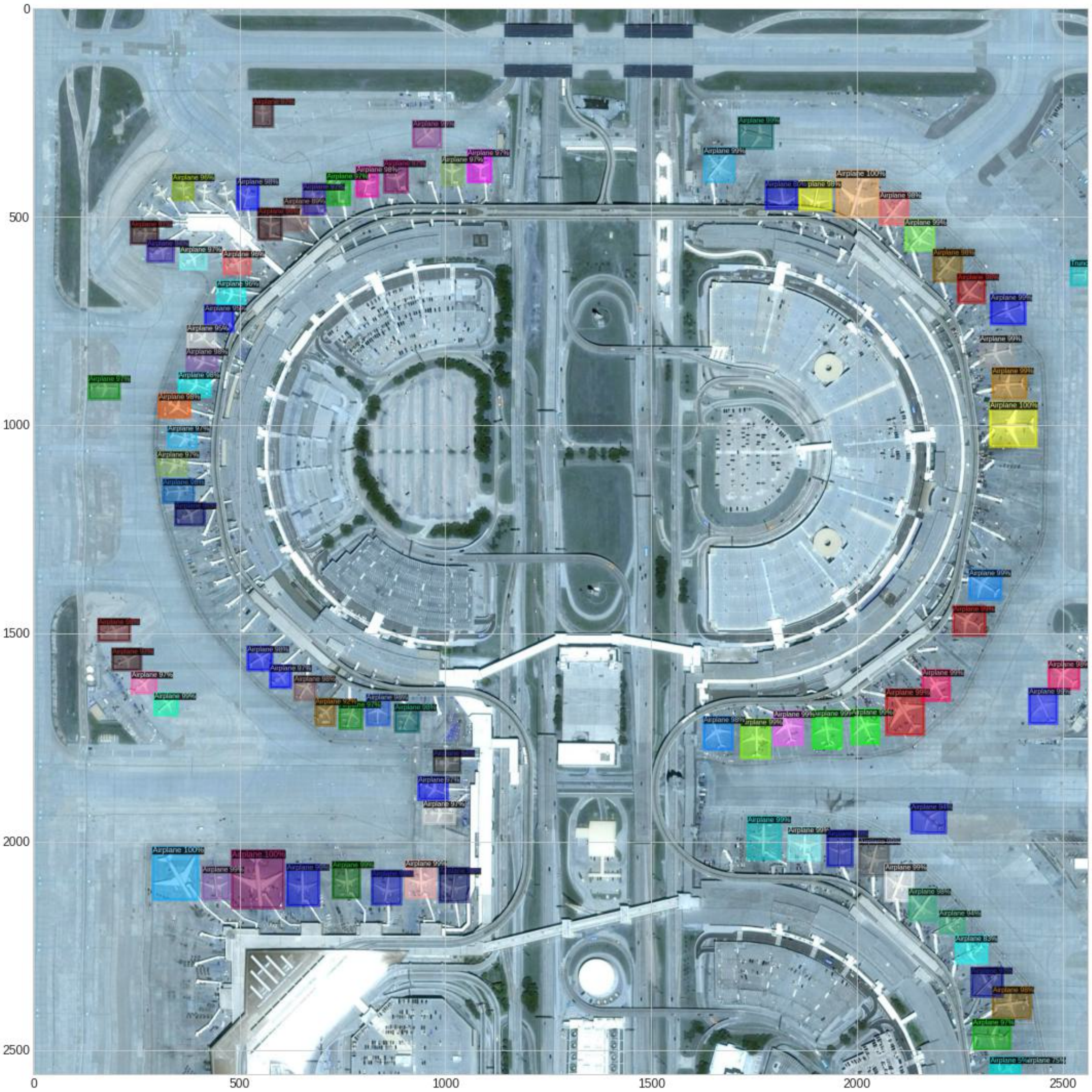

**Fig. 17 Example of detection results on the Airbus Aircraft dataset with ViTDet, ViT-L backbone.**

### 5.4 RarePlanes Object Detection Results

The experiment steps for RarePlanes are the same as the ones used on the Airbus Aircraft dataset, except for dataset split because RarePlanes already provides training and testing sub-datasets. Table 7 lists the information of the provided training and testing sub-datasets. We ran experiments for two types of object detection tasks: aircraft and aircraft role. Next, the detailed results on four experiments will be presented.

**Table 7 Training and testing datasets of RarePlanes**

| | Real | Synthetic |
|---|---|---|
| Training | 5,815 | 45,000 |
| Testing | 2,710 | 5,000 |
| Total | 8,525 | 50,000 |

#### *5.4.1. Aircraft Object Detection Results for the Real Image Dataset*

Table 8 shows aircraft object detection results for the real dataset. The best metrics across algorithms are highlighted in yellow. Like the previous findings, Cascade Mask RCNN still outperforms Mask RCNN, and ViTDet still beats CNN backbone in all evaluation metrics. For small object detection, ViTDet can beat the CNN counterpart by 7-11% on APs, which implies ViTDet backbone can better detect small objects. Fig. 18 shows an example of object detection results on a testing image. The two aircraft are tested with high confidence value (>=98%).

#### *5.4.2. Aircraft Object Detection Results for the Synthetic Image Dataset*

Table 9 presents aircraft object detection results for the synthetic dataset. As with the previous testing results, Cascade Mask RCNN still consistently outperforms Mask RCNN; ViTDet still beats CNN backbone in all evaluation metrics. For this dataset, the performance improvement of small object detection is not so large as in the two previously tested datasets. Fig. 19 shows an example of aircraft object detection on a testing image. In this case, all aircraft are identified. However, several non-aircraft objects are mis-labeled as aircraft.

#### *5.4.3. Aircraft Role Object Detection Results for the Real Image Dataset*

Table 10 shows aircraft role object detection results for the real dataset. As with the above three experiment cases, Cascade Mask RCNN still outperforms Mask RCNN, except on AP50. For AP, ViTDet backbone still beats CNN backbone with large margins of improvement (14-17%). Fig. 20 shows an example of aircraft role object detection on a testing image, where the aircraft are labeled by their role names of "large civil transportation utility."

#### *5.4.4. Aircraft Role Object Detection Results for the Synthetic Image Dataset*

Table 11 presents aircraft role object detection results for the synthetic dataset. Similarly, Cascade Mask RCNN still performs better than Mask RCNN. On AP, ViTDet backbone still beats CNN backbone with large margins (12-16%). Fig. 21 shows an example of aircraft role object detection on a testing image, and roles are identified according to their sizes. As in Fig. 19, all aircraft objects are correctly identified with their roles. However, several non-aircraft objects are wrongly labeled as aircraft role.

As seen in the above four experiments for the RarePlanes dataset, obviously ViTDet backbone performs much better than the CNN counterpart. For AP, the improvement ranges from 5% to 17%. The above experiments focused on HBB object detection performed with Detectron2. Next, we will present OBB object detection with MMRotate.

**Table 8 RarePlanes real dataset aircraft object detection results comparison**

| Method | Backbone | Pre-train | Task | FT-epoch | Learning rate | AP | AP50 | AP75 | APs | APm | APl |
|---|---|---|---|---|---|---|---|---|---|---|---|
| Mask RCNN | ResNeXt-101 | 1K,sup | aircarft | 1000 | 0.0001 | 69.17 | 96.33 | 85.78 | 58.66 | 68.57 | 83.86 |
| Mask RCNN | ViTDet,ViT-L | 1K,MAE | aircarft | 1000 | 0.0001 | 74.72 | 98.29 | 88.62 | 65.55 | 73.22 | 84.78 |
| Cascade Mask RCNN | ViTDet,ViT-L | 1K,MAE | aircarft | 1000 | 0.0001 | 77.45 | 97.60 | 89.97 | 70.02 | 75.81 | 87.17 |

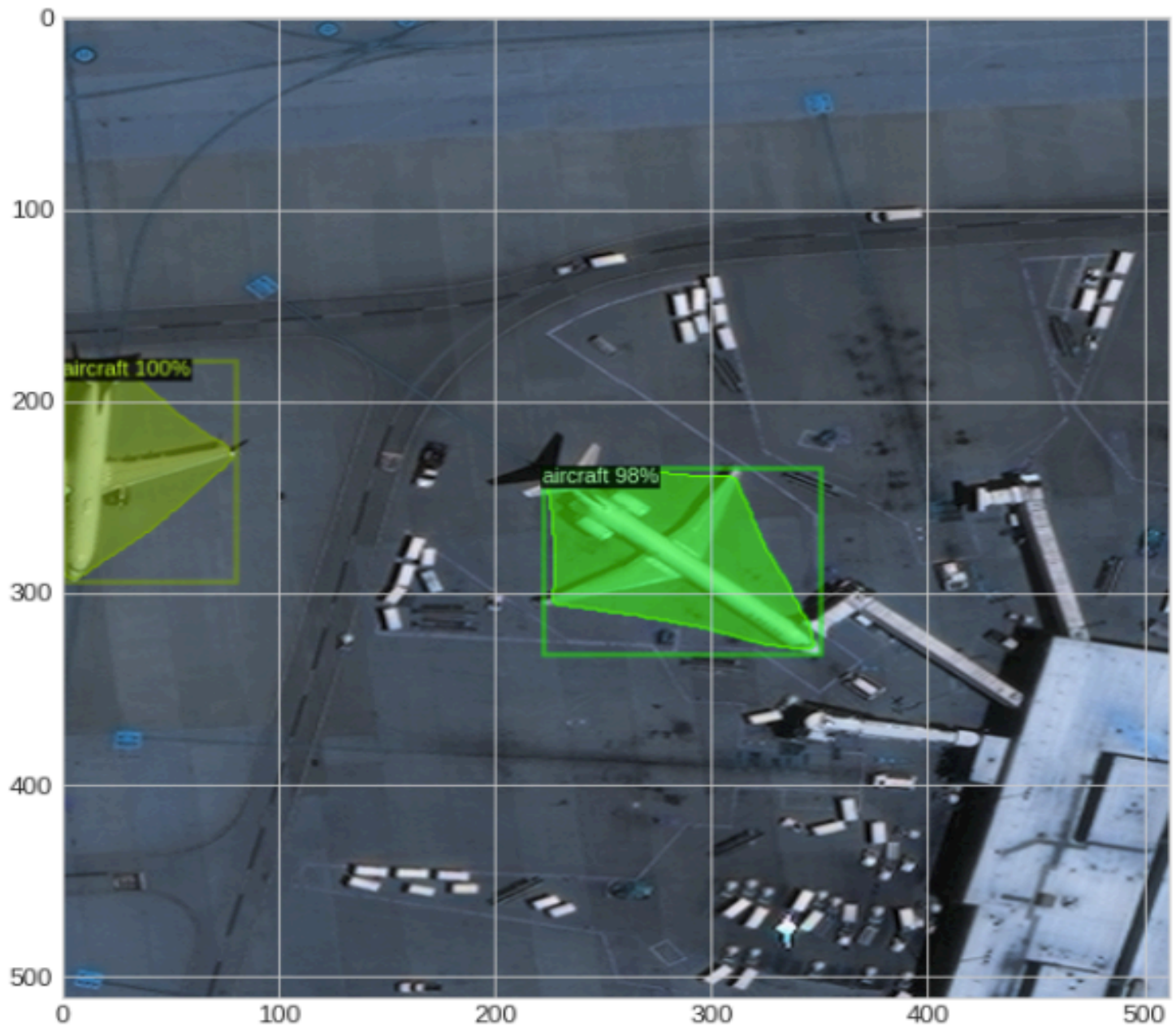

Fig. 18 Example of aircraft object detection results on the RarePlanes real testing dataset.

Table 9 RarePlanes synthetic dataset aircraft object detection results comparison

| Method | Backbone | Pre-train | Task | FT-epoch | Learning rate | AP | AP50 | AP75 | APs | APm | APl |
|---|---|---|---|---|---|---|---|---|---|---|---|
| Mask RCNN | ResNeXt-101 | 1K,sup | aircarft | 1000 | 0.0001 | 69.43 | 91.89 | 84.66 | 39.91 | 67.13 | 81.02 |
| Mask RCNN | ViTDet,ViT-L | 1K,MAE | aircarft | 1000 | 0.0001 | 74.80 | 96.69 | 84.85 | 40.13 | 69.32 | 88.42 |
| Cascade Mask RCNN | ViTDet,ViT-L | 1K,MAE | aircarft | 1000 | 0.0001 | 78.06 | 96.70 | 87.17 | 43.33 | 73.06 | 91.58 |

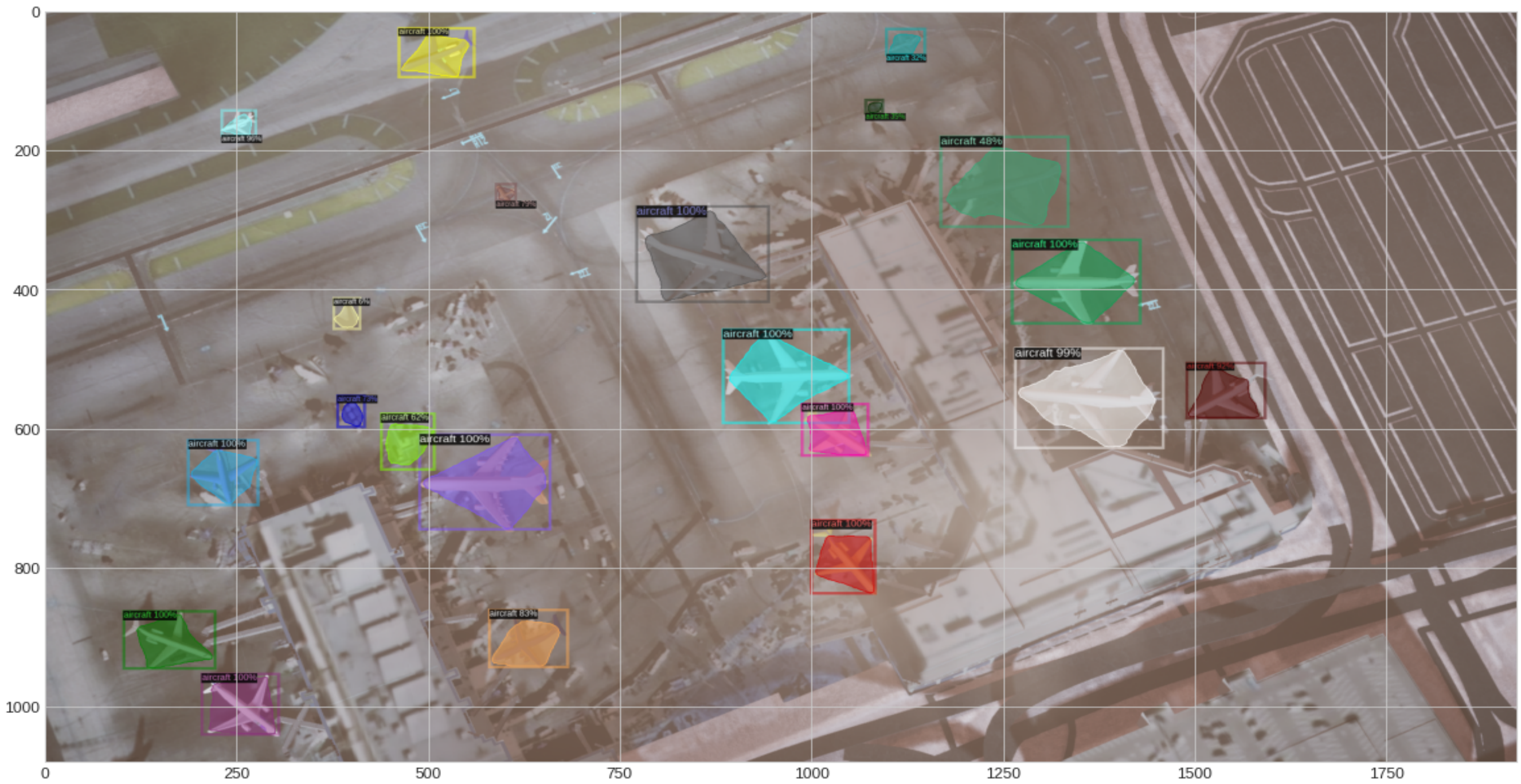

Fig. 19 Example of aircraft object detection results on the RarePlanes synthetic testing dataset.

**Table 10 RarePlanes real dataset aircraft role object detection result comparison**

| Method | Backbone | Pre-train | Task | FT-epoch | Learning rate | AP | AP50 | AP75 | APs | APm | APl |
|---|---|---|---|---|---|---|---|---|---|---|---|
| Mask RCNN | ResNeXt-101 | 1K,sup | aircarft role | 1000 | 0.0001 | 56.96 | 82.58 | 71.35 | 36.28 | 44.23 | 60.45 |
| Mask RCNN | ViTDet,ViT-L | 1K,MAE | aircarft role | 1000 | 0.0001 | 71.00 | 93.41 | 84.29 | 45.51 | 56.29 | 81.03 |
| Cascade Mask RCNN | ViTDet,ViT-L | 1K,MAE | aircarft role | 1000 | 0.0001 | 73.94 | 92.38 | 85.74 | 52.88 | 59.26 | 83.04 |

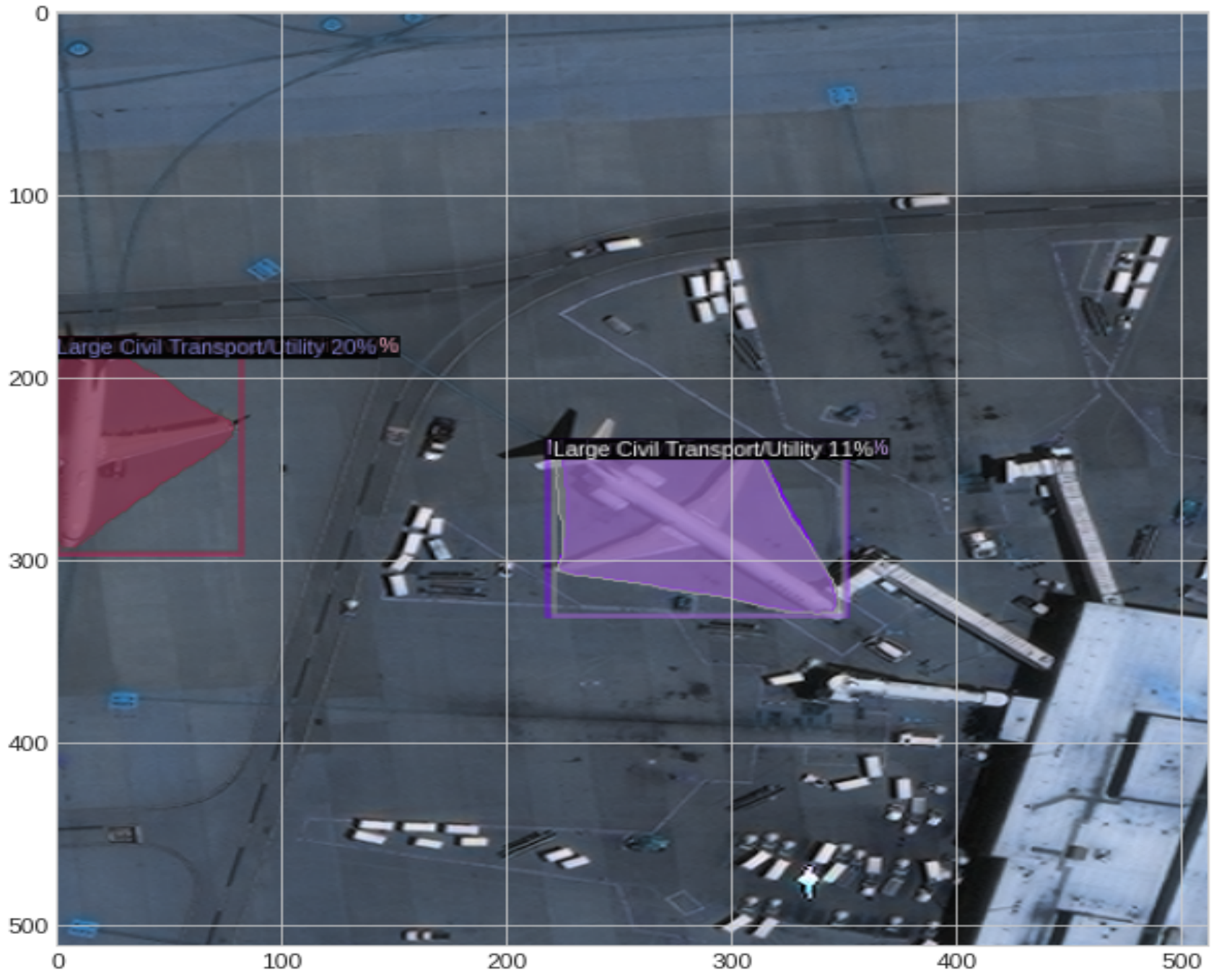

**Fig. 20 Example of aircraft role object detection results on the RarePlanes real dataset.**

**Table 11 RarePlanes synthetic dataset aircraft role object detection results comparison**

| Method | Backbone | Pre-train | Task | FT-epoch | Learning rate | AP | AP50 | AP75 | APs | APm | APl |
|---|---|---|---|---|---|---|---|---|---|---|---|
| Mask RCNN | ResNeXt-101 | 1K,sup | aircarft role | 1000 | 0.0001 | 52.66 | 71.31 | 66.11 | 14.23 | 39.32 | 40.78 |
| Mask RCNN | ViTDet,ViT-L | 1K,MAE | aircarft role | 1000 | 0.0001 | 64.62 | 90.93 | 70.56 | 23.46 | 56.17 | 54.02 |
| Cascade Mask RCNN | ViTDet,ViT-L | 1K,MAE | aircarft role | 1000 | 0.0001 | 68.67 | 91.78 | 75.32 | 27.56 | 60.88 | 57.27 |

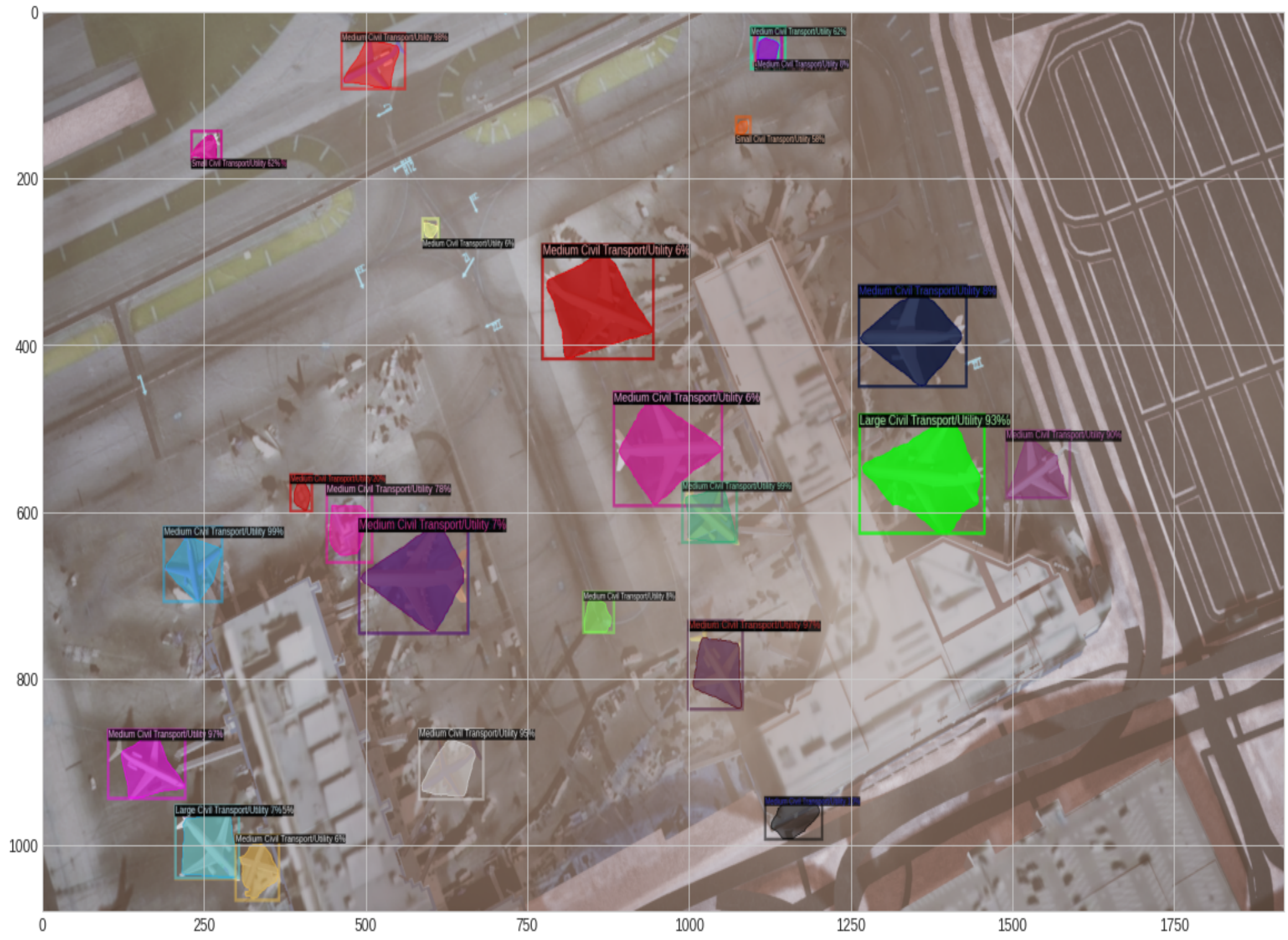

**Fig. 21 Example of aircraft role object detection results for RarePlanes synthetic dataset.**

### 5.5 DOTA-v1.0 Object Detection Results

Training models for DOTA-v1.0 is more complicated than the previous HBB object detection experiments. We followed five steps to carry out OBB experiments:

**Step 1. Data preprocessing**

For a fair comparison, we followed the same data preprocessing steps laid out in Ding et al. [1] and Han et al. [161]. Specifically, we first combined both training and validation sub-datasets to train models, and the testing dataset was used for final evaluation. Note that the testing dataset does not provide data labels in the downloaded folders, and the DOTA web evaluation server [172] must be used for the final results. DOTA's image size ranges from 800 x 800 to 4,000 x 4,000; therefore, we also followed the image splitting practice and cropped the original images into 1,024 × 1,024 patches with a stride of 824. Just as importantly, we also carried out data augmentation to get a variant of DOTA-v1.0 for training, in which we also adopted standard random rotation (RR) and multi-scale (MS) transformation at three scales {0.5, 1.0, and 1.5} for a fair comparison. After all the necessary steps, we have two transformed datasets for the model training; one is only with splitting, and the other is with splitting and data augmentation.

**Step 2. Pretrained models downloading**

From the RoI Transformer and ReDet model zoo, we downloaded the pretrained models with the ResNet and Swin-T backbones for comparing. For RoI Transformer, models were downloaded from the following webpage: https://github.com/open-mmlab/mmrotate/blob/main/configs/roi_trans/README.md . For ReDet, models were downloaded from the DOTA-v.10 table on the following webpage: https://github.com/open-mmlab/mmrotate/blob/main/configs/redet/README.md.

**Step 3. Configuration files customization**

We set up configuration files for ViTDet backbone in the selected algorithms: RoI Transformer [1] and ReDet [161]. We used the default configuration files provided in the MMRotate platform as exemplar and created corresponding ones for ViTDet backbone. In details, the angle representation was set to 1e90, the learning rate was 0.0001, AdamW optimizer was used, and the number of training epochs was 12.

**Step 4. Fine-tuning models for ViTDet backbone**

We fine-tuned four models with ViTDet backbone in RoI Transformer and ReDet on two preprocessed datasets in Step 1.

**Step 5. Evaluation on the testing dataset**

When all 9 models were ready, we evaluated them on the testing datasets. We then submitted the predicated object detection results to the DOTA-v1.0 official evaluation server, which in turn gave us AP for each class and mAP for all classes shown in Table 12.

Table 12 presents the detailed nine experiment results. The three backbones are as follows: R50 stands for ResNet-50; Swin-T represents Swin Transformer tiny version; and ViTDet, ViT-B is ViTDet backbone built with a base version of ViT that has 12 layers and 768-dimension output. With consideration of OBB algorithms' heavy computation burden, we did not evaluate ViTDet, ViT-L and ViTDet, ViT-H. The column "aug." shows whether data augmentation was used. As demonstrated in Table 12, given the same backbones, RetDet performs slightly better than RoI Transformer. Without data augmentation, ViTDet, ViT-B backbone is slightly worse than the other two backbones. However, with data augmentation, ViTDet, ViT-B achieves the best performance on mAP (80.89%), which is very comparable to the best published benchmark of 80.90% in Zhou et al. [169]. That benchmark was achieved with a combination of RoI Transformer, Swin-T backbone, Kullback-Leibler Divergence (KLD) trick [179], and data augmentation. In comparison, ViTDet can much more easily achieve comparable best performance without the need of the complicated KLD trick, in which the rotated bounding box must be converted into a 2-D Gaussian distribution and then KLD between the Gaussian distributions are calculated as the regression loss. Moreover, for helicopter detection (HC), ViTDet, ViT-B performs the best, improving about 23% at large. Overall, for a complicated dataset like DOTA, data augmentation still plays a bigger role than backbones.

Fig. 22 shows an example of detection results. In a compacted parking lot like the one pictured, most of the vehicles are detected with high confidence values. In short, for OBB object detection, ViTDet, ViT-B still achieves comparable performance with other backbones; the computation burden of ViTDet is heavier than R50 and Swin-T backbones. Therefore, more research may be needed to improve ViTDet's performance for OBB detection.

**Table 12 Accuracy comparison of rotated object detection on DOTA-v1.0**

| method | backbone | pre-train | aug. | PL | BD | BR | GTF | SV | LV | SH | TC | BC | ST | SBF | RA | HA | SP | HC | mAP |
|---|---|---|---|---|---|---|---|---|---|---|---|---|---|---|---|---|---|---|---|
| RoI Trans | R50 | 1K,sup | - | 88.97 | 82.14 | 54.59 | 76.28 | 79.29 | 77.94 | 87.94 | 90.88 | 87.19 | 85.62 | 62.21 | 62.63 | 74.62 | 72.43 | 59.23 | 76.13 |
| RoI Trans | Swin-T | 1K,sup | - | 89.08 | 83.60 | 54.84 | 72.10 | 79.02 | 84.45 | 87.97 | 90.90 | 87.14 | 86.64 | 64.65 | 66.50 | 76.65 | 72.30 | 66.90 | 77.52 |
| RoI Trans | ViTDet,ViT-B | 1K,MAE | - | 89.42 | 81.13 | 52.99 | 72.25 | 78.35 | 84.26 | 88.16 | 90.89 | 84.58 | 86.63 | 53.09 | 66.37 | 75.37 | 72.63 | 59.20 | 75.69 |
| RoI Trans | R50 | 1K,sup | MS+RR | 88.76 | 84.47 | 59.20 | 78.65 | 79.65 | 85.50 | 88.26 | 90.90 | 87.05 | 88.21 | 69.73 | 68.77 | 78.90 | 81.48 | 71.36 | 80.06 |
| RoI Trans | ViTDet,ViT-B | 1K,MAE | MS+RR | 88.87 | 84.51 | 60.31 | 75.48 | 81.04 | 86.00 | 88.47 | 90.84 | 84.63 | 87.62 | 62.81 | 72.07 | 78.86 | 82.47 | 75.94 | 79.99 |
| ReDet | R50 | 1K,sup | - | 89.20 | 83.79 | 52.23 | 73.31 | 78.06 | 82.48 | 88.23 | 90.86 | 87.26 | 85.97 | 65.64 | 62.87 | 75.90 | 70.02 | 66.79 | 76.84 |
| ReDet | R50 | 1K,sup | MS+RR | 89.19 | 85.75 | 62.13 | 81.20 | 78.98 | 86.01 | 88.67 | 90.90 | 89.20 | 88.23 | 69.81 | 66.54 | 79.13 | 78.72 | 71.19 | 80.38 |
| ReDet | ViTDet,ViT-B | 1K,MAE | - | 89.24 | 80.37 | 53.84 | 71.37 | 78.21 | 84.19 | 88.12 | 90.90 | 85.51 | 86.28 | 52.20 | 65.94 | 75.91 | 70.93 | 62.63 | 75.71 |
| ReDet | ViTDet,ViT-B | 1K,MAE | MS+RR | 87.75 | 85.22 | 61.37 | 81.12 | 80.62 | 85.82 | 88.37 | 90.88 | 85.93 | 87.79 | 63.31 | 73.15 | 78.96 | 80.13 | 82.88 | 80.89 |

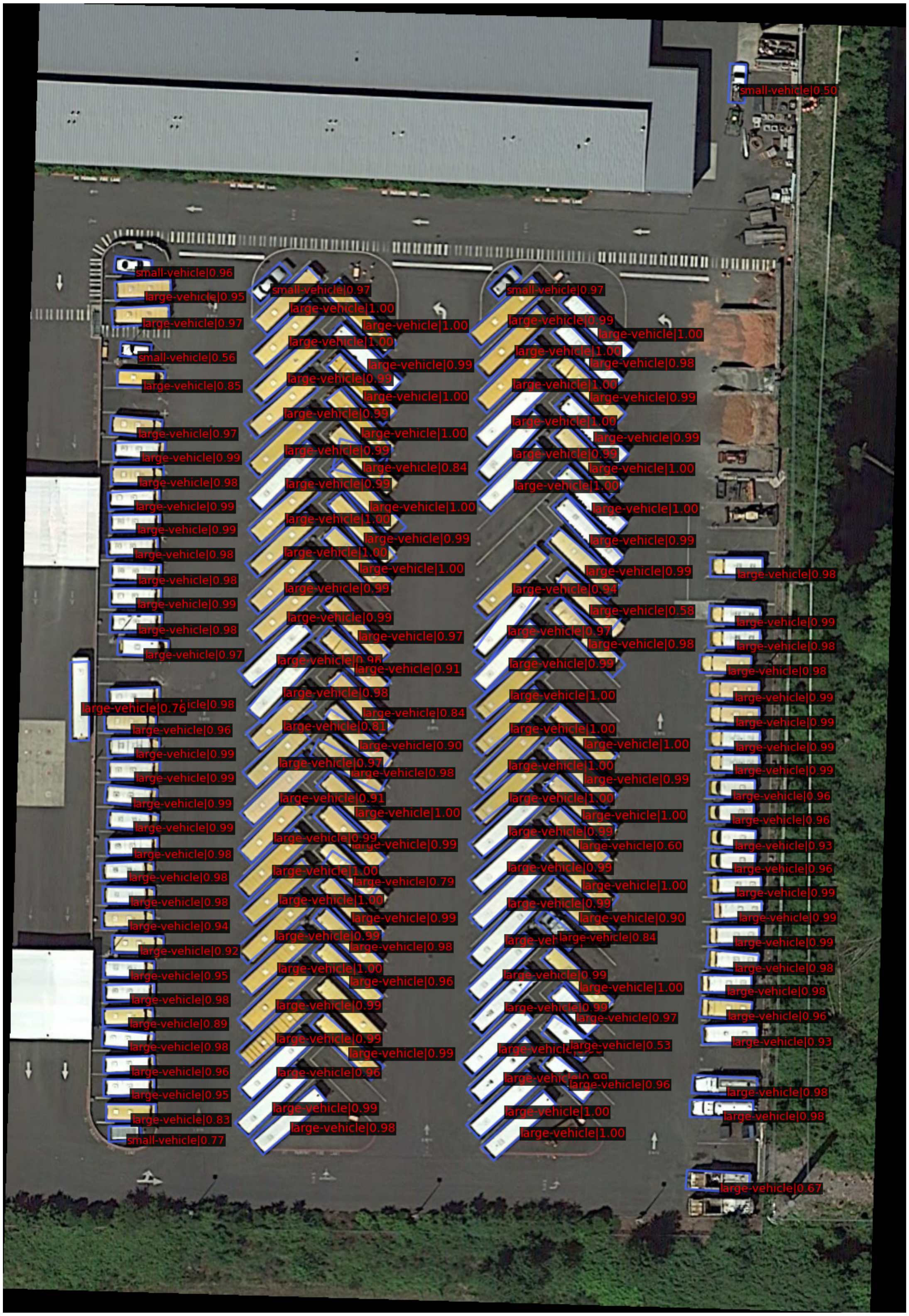

**Fig. 22 Example of object detection results on DOTA-v1.0 with ViTDet backbone.**

## 6. CONCLUSION

This study has explored the newly proposed MAE pretrained ViTDet backbone for challenging aerial image object detection problems. We carried out 25 experiments on three well-known aerial image datasets: Airbus Aircraft, RarePlanes, and DOTA. Our experiments demonstrated that ViTDet backbone consistently beats its CNN counterparts in HBB object detection (up to 17% improvement on AP) and that it achieves on-par performance for OBB object detection. Our results also provided a baseline for future research.

## ACKNOWLEDGMENTS

The authors sincerely thank Dr. Kris Rosfjord and Dr. Heath Farris for their generous support of this project. We would also like to thank Mike Robinson, Bill Bateman, Lixia Song, Erik Vargo, and Paul A. Diffenderfer of The MITRE Corporation for their valuable discussions, insights, and encouragement.

## NOTICE

This work was sponsored by MITRE's independent research and development program. The contents of this document reflect the views of the authors and do not necessarily reflect the views of the Federal Aviation Administration (FAA) or the Department of Transportation (DOT). Neither the FAA nor the DOT makes any warranty or guarantee, expressed or implied, concerning the content or accuracy of these views.

## REFERENCES

[1] J. Ding, N. Xue, Y. Long, G.-S. Xia, and Q. Lu, "Learning RoI Transformer for Detecting Oriented Objects in Aerial Images," Dec. 2018, doi: 10.48550/arXiv.1812.00155.

[2] K. He, X. Zhang, S. Ren, and J. Sun, "Deep Residual Learning for Image Recognition." arXiv, Dec. 10, 2015. Accessed: Dec. 01, 2022. [Online]. Available: http://arxiv.org/abs/1512.03385

[3] A. Dosovitskiy *et al.*, "An Image is Worth 16x16 Words: Transformers for Image Recognition at Scale." arXiv, Jun. 03, 2021. doi: 10.48550/arXiv.2010.11929.

[4] Z. Liu *et al.*, "Swin Transformer: Hierarchical Vision Transformer using Shifted Windows." arXiv, Aug. 17, 2021. Accessed: Dec. 01, 2022. [Online]. Available: http://arxiv.org/abs/2103.14030

[5] Y. Li, H. Mao, R. Girshick, and K. He, "Exploring Plain Vision Transformer Backbones for Object Detection." arXiv, Jun. 10, 2022. Accessed: Nov. 29, 2022. [Online]. Available: http://arxiv.org/abs/2203.16527

[6] T.-Y. Lin, P. Dollár, R. Girshick, K. He, B. Hariharan, and S. Belongie, "Feature Pyramid Networks for Object Detection." arXiv, Apr. 19, 2017. Accessed: Dec. 01, 2022. [Online]. Available: http://arxiv.org/abs/1612.03144

[7] K. He, X. Chen, S. Xie, Y. Li, P. Dollár, and R. Girshick, "Masked Autoencoders Are Scalable Vision Learners." arXiv, Dec. 19, 2021. doi: 10.48550/arXiv.2111.06377.

[8] Synced, "Kaiming He's MetaAI Team Proposes ViTDet: A Plain Vision Transformer Backbone Competitive With…," *SyncedReview*, Apr. 07, 2022. https://medium.com/syncedreview/kaiming-hes-metaai-team-proposes-vitdet-a-plain-vision-transformer-backbone-competitive-with-ff4ad0814243 (accessed Dec. 03, 2022).

[9] "COCO - Common Objects in Context." https://cocodataset.org/#home (accessed Nov. 30, 2022).

[10] H. Bao, L. Dong, S. Piao, and F. Wei, "BEiT: BERT Pre-Training of Image Transformers." arXiv, Sep. 03, 2022. doi: 10.48550/arXiv.2106.08254.

[11] Z. Peng, L. Dong, H. Bao, Q. Ye, and F. Wei, "BEiT v2: Masked Image Modeling with Vector-Quantized Visual Tokenizers." arXiv, Oct. 03, 2022. Accessed: Nov. 30, 2022. [Online]. Available: http://arxiv.org/abs/2208.06366

[12] Z. Xie *et al.*, "SimMIM: A Simple Framework for Masked Image Modeling." arXiv, Apr. 17, 2022. doi: 10.48550/arXiv.2111.09886.

[13] Y. Shi, N. Siddharth, P. H. S. Torr, and A. R. Kosiorek, "Adversarial Masking for Self-Supervised Learning." arXiv, Jul. 06, 2022. doi: 10.48550/arXiv.2201.13100.

[14] J. Gui, Z. Liu, and H. Luo, "Good helper is around you: Attention-driven Masked Image Modeling." arXiv, Nov. 28, 2022. Accessed: Nov. 30, 2022. [Online]. Available: http://arxiv.org/abs/2211.15362

[15] I. Kakogeorgiou *et al.*, "What to Hide from Your Students: Attention-Guided Masked Image Modeling," Mar. 2022, doi: 10.1007/978-3-031-20056-4_18.

[16] Y. Tian *et al.*, "Beyond Masking: Demystifying Token-Based Pre-Training for Vision Transformers." arXiv, Mar. 27, 2022. Accessed: Nov. 29, 2022. [Online]. Available: http://arxiv.org/abs/2203.14313

[17] X. Dong *et al.*, "Bootstrapped Masked Autoencoders for Vision BERT Pretraining." arXiv, Jul. 14, 2022. Accessed: Nov. 30, 2022. [Online]. Available: http://arxiv.org/abs/2207.07116

[18] X. Chen *et al.*, "Context Autoencoder for Self-Supervised Representation Learning." arXiv, May 30, 2022. doi: 10.48550/arXiv.2202.03026.

[19] Anonymous, "CAN: A simple, efficient and scalable contrastive masked autoencoder framework for learning visual representations," presented at the The Eleventh International Conference on Learning Representations, Nov. 2022. Accessed: Nov. 29, 2022. [Online]. Available: https://openreview.net/forum?id=qmV_tOHp7B9

[20] P. Gao, T. Ma, H. Li, Z. Lin, J. Dai, and Y. Qiao, "ConvMAE: Masked Convolution Meets Masked Autoencoders." arXiv, May 19, 2022. doi: 10.48550/arXiv.2205.03892.

[21] Z. Huang *et al.*, "Contrastive Masked Autoencoders are Stronger Vision Learners." arXiv, Nov. 28, 2022. doi: 10.48550/arXiv.2207.13532.

[22] X. Wang, K. Zhao, R. Zhang, S. Ding, Y. Wang, and W. Shen, "ContrastMask: Contrastive Learning to Segment Every Thing." arXiv, Mar. 24, 2022. doi: 10.48550/arXiv.2203.09775.

[23] X. Liu, J. Zhou, T. Kong, X. Lin, and R. Ji, "Exploring Target Representations for Masked Autoencoders." arXiv, Nov. 03, 2022. Accessed: Nov. 30, 2022. [Online]. Available: http://arxiv.org/abs/2209.03917

[24] Y. Bai *et al.*, "Masked Autoencoders Enable Efficient Knowledge Distillers." arXiv, Nov. 09, 2022. Accessed: Nov. 30, 2022. [Online]. Available: http://arxiv.org/abs/2208.12256

[25] Q. Wu, H. Ye, Y. Gu, H. Zhang, L. Wang, and D. He, "Denoising Masked Autoencoders are Certifiable Robust Vision Learners." arXiv, Nov. 01, 2022. Accessed: Nov. 30, 2022. [Online]. Available: http://arxiv.org/abs/2210.06983

[26] L. Huang, S. You, M. Zheng, F. Wang, C. Qian, and T. Yamasaki, "Green Hierarchical Vision Transformer for Masked Image Modeling." arXiv, Oct. 14, 2022. Accessed: Nov. 29, 2022. [Online]. Available: http://arxiv.org/abs/2205.13515

[27] J. Zhou *et al.*, "iBOT: Image BERT Pre-Training with Online Tokenizer." arXiv, Jan. 27, 2022. Accessed: Nov. 29, 2022. [Online]. Available: http://arxiv.org/abs/2111.07832

[28] J. Chen, M. Hu, B. Li, and M. Elhoseiny, "Efficient Self-supervised Vision Pretraining with Local Masked Reconstruction." arXiv, Jun. 20, 2022. doi: 10.48550/arXiv.2206.00790.

[29] R. Hu, S. Debnath, S. Xie, and X. Chen, "Exploring Long-Sequence Masked Autoencoders." arXiv, Oct. 13, 2022. Accessed: Nov. 30, 2022. [Online]. Available: http://arxiv.org/abs/2210.07224

[30] H. Xue *et al.*, "Stare at What You See: Masked Image Modeling without Reconstruction." arXiv, Nov. 16, 2022. Accessed: Nov. 30, 2022. [Online]. Available: http://arxiv.org/abs/2211.08887

[31] Z. Peng, L. Dong, H. Bao, Q. Ye, and F. Wei, "A Unified View of Masked Image Modeling." arXiv, Oct. 19, 2022. Accessed: Nov. 30, 2022. [Online]. Available: http://arxiv.org/abs/2210.10615

[32] C. Wei, H. Fan, S. Xie, C.-Y. Wu, A. Yuille, and C. Feichtenhofer, "Masked Feature Prediction for Self-Supervised Visual Pre-Training." arXiv, Dec. 16, 2021. doi: 10.48550/arXiv.2112.09133.

[33] S. A. Taghanaki *et al.*, "MaskTune: Mitigating Spurious Correlations by Forcing to Explore." arXiv, Oct. 08, 2022. Accessed: Nov. 30, 2022. [Online]. Available: http://arxiv.org/abs/2210.00055

[34] J. Li *et al.*, "MetaMask: Revisiting Dimensional Confounder for Self-Supervised Learning." arXiv, Sep. 16, 2022. Accessed: Nov. 30, 2022. [Online]. Available: http://arxiv.org/abs/2209.07902

[35] J. Xie, W. Li, X. Zhan, Z. Liu, Y. S. Ong, and C. C. Loy, "Masked Frequency Modeling for Self-Supervised Visual Pre-Training." arXiv, Jun. 15, 2022. Accessed: Nov. 30, 2022. [Online]. Available: http://arxiv.org/abs/2206.07706

[36] Z. Hou, F. Sun, Y.-K. Chen, Y. Xie, and S.-Y. Kung, "MILAN: Masked Image Pretraining on Language Assisted Representation." arXiv, Aug. 15, 2022. Accessed: Nov. 30, 2022. [Online]. Available: http://arxiv.org/abs/2208.06049

[37] K. Vishniakov, E. Xing, and Z. Shen, "MixMask: Revisiting Masked Siamese Self-supervised Learning in Asymmetric Distance." arXiv, Nov. 24, 2022. doi: 10.48550/arXiv.2210.11456.

[38] J. Liu, X. Huang, O. Yoshie, Y. Liu, and H. Li, "MixMIM: Mixed and Masked Image Modeling for Efficient Visual Representation Learning." arXiv, Sep. 06, 2022. Accessed: Nov. 29, 2022. [Online]. Available: http://arxiv.org/abs/2205.13137

[39] H. Xu, S. Ding, X. Zhang, H. Xiong, and Q. Tian, "Masked Autoencoders are Robust Data Augmentors." arXiv, Jun. 09, 2022. Accessed: Nov. 30, 2022. [Online]. Available: http://arxiv.org/abs/2206.04846

[40] M. Assran *et al.*, "Masked Siamese Networks for Label-Efficient Learning." arXiv, Apr. 14, 2022. Accessed: Nov. 29, 2022. [Online]. Available: http://arxiv.org/abs/2204.07141

[41] Z. Li *et al.*, "MST: Masked Self-Supervised Transformer for Visual Representation." arXiv, Oct. 24, 2021. Accessed: Nov. 29, 2022. [Online]. Available: http://arxiv.org/abs/2106.05656

[42] R. Bachmann, D. Mizrahi, A. Atanov, and A. Zamir, "MultiMAE: Multi-modal Multi-task Masked Autoencoders." arXiv, Apr. 04, 2022. doi: 10.48550/arXiv.2204.01678.

[43] L. Wei, L. Xie, W. Zhou, H. Li, and Q. Tian, "MVP: Multimodality-guided Visual Pre-training." arXiv, Mar. 10, 2022. doi: 10.48550/arXiv.2203.05175.

[44] Y. Lee, J. Willette, J. Kim, J. Lee, and S. J. Hwang, "Exploring The Role of Mean Teachers in Self-supervised Masked Auto-Encoders." arXiv, Oct. 05, 2022. Accessed: Nov. 30, 2022. [Online]. Available: http://arxiv.org/abs/2210.02077

[45] S. Lee, M. Kang, J. Lee, S. J. Hwang, and K. Kawaguchi, "Self-Distillation for Further Pre-training of Transformers." arXiv, Sep. 29, 2022. Accessed: Nov. 30, 2022. [Online]. Available: http://arxiv.org/abs/2210.02871

[46] G. Li, H. Zheng, D. Liu, C. Wang, B. Su, and C. Zheng, "SemMAE: Semantic-Guided Masking for Learning Masked Autoencoders." arXiv, Oct. 05, 2022. Accessed: Nov. 30, 2022. [Online]. Available: http://arxiv.org/abs/2206.10207

[47] Y. Chen *et al.*, "SdAE: Self-distillated Masked Autoencoder." arXiv, Jul. 31, 2022. Accessed: Nov. 29, 2022. [Online]. Available: http://arxiv.org/abs/2208.00449

[48] F. Liang, Y. Li, and D. Marculescu, "SupMAE: Supervised Masked Autoencoders Are Efficient Vision Learners," May 2022, doi: 10.48550/arXiv.2205.14540.

[49] Q. Zhang, Y. Wang, and Y. Wang, "How Mask Matters: Towards Theoretical Understandings of Masked Autoencoders." arXiv, Oct. 15, 2022. Accessed: Nov. 30, 2022. [Online]. Available: http://arxiv.org/abs/2210.08344

[50] X. Li, W. Wang, L. Yang, and J. Yang, "Uniform Masking: Enabling MAE Pre-training for Pyramid-based Vision Transformers with Locality," May 2022, doi: 10.48550/arXiv.2205.10063.

[51] W. G. C. Bandara, N. Patel, A. Gholami, M. Nikkhah, M. Agrawal, and V. M. Patel, "AdaMAE: Adaptive Masking for Efficient Spatiotemporal Learning with Masked Autoencoders." arXiv, Nov. 16, 2022. Accessed: Nov. 30, 2022. [Online]. Available: http://arxiv.org/abs/2211.09120

[52] R. Wang *et al.*, "BEVT: BERT Pretraining of Video Transformers." arXiv, Mar. 03, 2022. Accessed: Nov. 29, 2022. [Online]. Available: http://arxiv.org/abs/2112.01529

[53] Y. Song, M. Yang, W. Wu, D. He, F. Li, and J. Wang, "It Takes Two: Masked Appearance-Motion Modeling for Self-supervised Video Transformer Pre-training." arXiv, Oct. 11, 2022. Accessed: Nov. 30, 2022. [Online]. Available: http://arxiv.org/abs/2210.05234

[54] Z. Qing *et al.*, "MAR: Masked Autoencoders for Efficient Action Recognition." arXiv, Jul. 24, 2022. Accessed: Nov. 29, 2022. [Online]. Available: http://arxiv.org/abs/2207.11660

[55] A. Gupta, S. Tian, Y. Zhang, J. Wu, R. Martín-Martín, and L. Fei-Fei, "MaskViT: Masked Visual Pre-Training for Video Prediction." arXiv, Aug. 06, 2022. Accessed: Nov. 29, 2022. [Online]. Available: http://arxiv.org/abs/2206.11894

[56] X. Sun, P. Chen, L. Chen, T. H. Li, M. Tan, and C. Gan, "M$^3$Video: Masked Motion Modeling for Self-Supervised Video Representation Learning." arXiv, Oct. 12, 2022. Accessed: Nov. 30, 2022. [Online]. Available: http://arxiv.org/abs/2210.06096

[57] V. Voleti, A. Jolicoeur-Martineau, and C. Pal, "MCVD: Masked Conditional Video Diffusion for Prediction, Generation, and Interpolation," May 2022, doi: 10.48550/arXiv.2205.09853.

[58] H. Yang *et al.*, "Self-supervised Video Representation Learning with Motion-Aware Masked Autoencoders." arXiv, Oct. 08, 2022. Accessed: Nov. 30, 2022. [Online]. Available: http://arxiv.org/abs/2210.04154

[59] R. Girdhar, A. El-Nouby, M. Singh, K. V. Alwala, A. Joulin, and I. Misra, "OmniMAE: Single Model Masked Pretraining on Images and Videos." arXiv, Jun. 16, 2022. Accessed: Nov. 29, 2022. [Online]. Available: http://arxiv.org/abs/2206.08356

[60] C. Feichtenhofer, H. Fan, Y. Li, and K. He, "Masked Autoencoders As Spatiotemporal Learners." arXiv, Oct. 21, 2022. Accessed: Nov. 29, 2022. [Online]. Available: http://arxiv.org/abs/2205.09113

[61] Y. Wang, J. Wang, B. Chen, Z. Zeng, and S. Xia, "Contrastive Masked Autoencoders for Self-Supervised Video Hashing." arXiv, Nov. 23, 2022. Accessed: Nov. 30, 2022. [Online]. Available: http://arxiv.org/abs/2211.11210

[62] Z. Tong, Y. Song, J. Wang, and L. Wang, "VideoMAE: Masked Autoencoders are Data-Efficient Learners for Self-Supervised Video Pre-Training." arXiv, Oct. 18, 2022. Accessed: Nov. 29, 2022. [Online]. Available: http://arxiv.org/abs/2203.12602

[63] H. Tan, J. Lei, T. Wolf, and M. Bansal, "VIMPAC: Video Pre-Training via Masked Token Prediction and Contrastive Learning." arXiv, Jun. 21, 2021. Accessed: Nov. 29, 2022. [Online]. Available: http://arxiv.org/abs/2106.11250

[64] S. Hwang, J. Yoon, Y. Lee, and S. J. Hwang, "Efficient Video Representation Learning via Masked Video Modeling with Motion-centric Token Selection." arXiv, Nov. 19, 2022. Accessed:

Nov. 30, 2022. [Online]. Available: http://arxiv.org/abs/2211.10636

[65] S. T. Ly, B. Lin, H. Q. Vo, D. Maric, B. Roysam, and H. V. Nguyen, "Student Collaboration Improves Self-Supervised Learning: Dual-Loss Adaptive Masked Autoencoder for Multiplexed Immunofluorescence Brain Images Analysis." arXiv, Aug. 18, 2022. Accessed: Nov. 29, 2022. [Online]. Available: http://arxiv.org/abs/2205.05194

[66] H. Quan *et al.*, "Global Contrast Masked Autoencoders Are Powerful Pathological Representation Learners." arXiv, May 21, 2022. Accessed: Nov. 29, 2022. [Online]. Available: http://arxiv.org/abs/2205.09048

[67] Y. Luo, Z. Chen, and X. Gao, "Self-distillation Augmented Masked Autoencoders for Histopathological Image Classification." arXiv, Jul. 25, 2022. Accessed: Nov. 29, 2022. [Online]. Available: http://arxiv.org/abs/2203.16983

[68] J. Jiang, N. Tyagi, K. Tringale, C. Crane, and H. Veeraraghavan, "Self-supervised 3D anatomy segmentation using self-distilled masked image transformer (SMIT)," May 2022, doi: 10.1007/978-3-031-16440-8_53.

[69] Y. Cong *et al.*, "SatMAE: Pre-training Transformers for Temporal and Multi-Spectral Satellite Imagery." arXiv, Oct. 19, 2022. Accessed: Nov. 29, 2022. [Online]. Available: http://arxiv.org/abs/2207.08051

[70] J. Li, S. Savarese, and S. C. H. Hoi, "Masked Unsupervised Self-training for Zero-shot Image Classification." arXiv, Jun. 06, 2022. doi: 10.48550/arXiv.2206.02967.

[71] X. Zhang *et al.*, "Integral Migrating Pre-trained Transformer Encoder-decoders for Visual Object Detection." arXiv, May 19, 2022. Accessed: Nov. 30, 2022. [Online]. Available: http://arxiv.org/abs/2205.09613

[72] F. Li *et al.*, "Mask DINO: Towards A Unified Transformer-based Framework for Object Detection and Segmentation." arXiv, Jun. 06, 2022. doi: 10.48550/arXiv.2206.02777.

[73] J. Wu and S. Mo, "Object-wise Masked Autoencoders for Fast Pre-training," May 2022, doi: 10.48550/arXiv.2205.14338.

[74] V. Prabhu, S. Yenamandra, A. Singh, and J. Hoffman, "Adapting Self-Supervised Vision Transformers by Probing Attention-Conditioned Masking Consistency." arXiv, Jun. 16, 2022. Accessed: Nov. 30, 2022. [Online]. Available: http://arxiv.org/abs/2206.08222

[75] Q. Yu *et al.*, "k-means Mask Transformer." arXiv, Oct. 31, 2022. Accessed: Nov. 30, 2022. [Online]. Available: http://arxiv.org/abs/2207.04044

[76] Z. Ding, J. Wang, and Z. Tu, "Open-Vocabulary Panoptic Segmentation with MaskCLIP." arXiv, Aug. 18, 2022. Accessed: Nov. 30, 2022. [Online]. Available: http://arxiv.org/abs/2208.08984

[77] W. Van Gansbeke, S. Vandenhende, and L. Van Gool, "Discovering Object Masks with Transformers for Unsupervised Semantic Segmentation." arXiv, Jun. 13, 2022. Accessed: Nov. 30, 2022. [Online]. Available: http://arxiv.org/abs/2206.06363

[78] L. Ke, M. Danelljan, X. Li, Y.-W. Tai, C.-K. Tang, and F. Yu, "Mask Transfiner for High-Quality Instance Segmentation." arXiv, Nov. 26, 2021. Accessed: Nov. 29, 2022. [Online]. Available: http://arxiv.org/abs/2111.13673

[79] A. Bielski and P. Favaro, "MOVE: Unsupervised Movable Object Segmentation and Detection." arXiv, Oct. 20, 2022. Accessed: Nov. 30, 2022. [Online]. Available: http://arxiv.org/abs/2210.07920

[80] G. Shin, W. Xie, and S. Albanie, "NamedMask: Distilling Segmenters from Complementary Foundation Models." arXiv, Sep. 22, 2022. Accessed: Nov. 30, 2022. [Online]. Available: http://arxiv.org/abs/2209.11228

[81] G. Couairon, J. Verbeek, H. Schwenk, and M. Cord, "DiffEdit: Diffusion-based semantic image editing with mask guidance." arXiv, Oct. 20, 2022. Accessed: Nov. 30, 2022. [Online]. Available: http://arxiv.org/abs/2210.11427

[82] T. Li, H. Chang, S. K. Mishra, H. Zhang, D. Katabi, and D. Krishnan, "MAGE: MAsked Generative Encoder to Unify Representation Learning and Image Synthesis." arXiv, Nov. 16, 2022. Accessed: Nov. 30, 2022. [Online]. Available: http://arxiv.org/abs/2211.09117

[83] H. Chang, H. Zhang, L. Jiang, C. Liu, and W. T. Freeman, "MaskGIT: Masked Generative Image Transformer," Feb. 2022, doi: 10.48550/arXiv.2202.04200.

[84] D. Lee, C. Kim, S. Kim, M. Cho, and W.-S. Han, "Draft-and-Revise: Effective Image Generation with Contextual RQ-Transformer." arXiv, Jun. 09, 2022. doi: 10.48550/arXiv.2206.04452.

[85] K. Wang *et al.*, "FaceMAE: Privacy-Preserving Face Recognition via Masked Autoencoders," May 2022, doi: 10.48550/arXiv.2205.11090.

[86] S. Hao, C. Chen, Z. Chen, and K.-Y. K. Wong, "A Unified Framework for Masked and Mask-Free Face Recognition via Feature Rectification." arXiv, Feb. 15, 2022. Accessed: Nov. 30, 2022. [Online]. Available: http://arxiv.org/abs/2202.07358

[87] M. R. Al-Sinan, A. F. Haneef, and H. Luqman, "Ensemble Learning using Transformers and Convolutional Networks for Masked Face Recognition." arXiv, Oct. 10, 2022. Accessed: Nov. 30, 2022. [Online]. Available: http://arxiv.org/abs/2210.04816

[88] P. Lyu *et al.*, "MaskOCR: Text Recognition with Masked Encoder-Decoder Pretraining." arXiv, Jun. 01, 2022. doi: 10.48550/arXiv.2206.00311.

[89] A. Baevski, W.-N. Hsu, Q. Xu, A. Babu, J. Gu, and M. Auli, "data2vec: A General Framework for Self-supervised Learning in Speech, Vision and Language." arXiv, Oct. 25, 2022. Accessed: Nov. 29, 2022. [Online]. Available: http://arxiv.org/abs/2202.03555

[90] X. Geng, H. Liu, L. Lee, D. Schuurmans, S. Levine, and P. Abbeel, "Multimodal Masked Autoencoders Learn Transferable Representations." arXiv, Oct. 21, 2022. Accessed: Nov. 29, 2022. [Online]. Available: http://arxiv.org/abs/2205.14204

[91] Z. Zhao, L. Guo, X. He, S. Shao, Z. Yuan, and J. Liu, "MAMO: Masked Multimodal Modeling for Fine-Grained Vision-Language Representation Learning." arXiv, Oct. 09, 2022. Accessed: Nov. 30, 2022. [Online]. Available: http://arxiv.org/abs/2210.04183

[92] X. Dong *et al.*, "MaskCLIP: Masked Self-Distillation Advances Contrastive Language-Image Pretraining." arXiv, Aug. 25, 2022. Accessed: Nov. 30, 2022. [Online]. Available: http://arxiv.org/abs/2208.12262

[93] G. Kwon, Z. Cai, A. Ravichandran, E. Bas, R. Bhotika, and S. Soatto, "Masked Vision and Language Modeling for Multi-modal Representation Learning." arXiv, Aug. 03, 2022. Accessed: Nov. 30, 2022. [Online]. Available: http://arxiv.org/abs/2208.02131

[94] Z. Chen *et al.*, "Multi-Modal Masked Autoencoders for Medical Vision-and-Language Pre-Training." arXiv, Sep. 15, 2022. Accessed: Nov. 30, 2022. [Online]. Available: http://arxiv.org/abs/2209.07098

[95] T. Arici *et al.*, "MLIM: Vision-and-Language Model Pre-training with Masked Language and Image Modeling," Sep. 2021, doi: 10.48550/arXiv.2109.12178.

[96] M. Shukor, G. Couairon, and M. Cord, "Efficient Vision-Language Pretraining with Visual Concepts and Hierarchical Alignment." arXiv, Oct. 05, 2022. Accessed: Nov. 30, 2022. [Online]. Available: http://arxiv.org/abs/2208.13628

[97] H. Bao, W. Wang, L. Dong, and F. Wei, "VL-BEiT: Generative Vision-Language Pretraining." arXiv, Sep. 03, 2022. Accessed: Nov. 29, 2022. [Online]. Available: http://arxiv.org/abs/2206.01127

[98] L. Gui, Q. Huang, A. Hauptmann, Y. Bisk, and J. Gao, "Training Vision-Language Transformers from Captions Alone." arXiv, May 18, 2022. Accessed: Nov. 30, 2022. [Online]. Available: http://arxiv.org/abs/2205.09256

[99] T.-J. Fu *et al.*, "An Empirical Study of End-to-End Video-Language Transformers with Masked Visual Modeling." arXiv, Sep. 04, 2022. Accessed: Nov. 29, 2022. [Online]. Available: http://arxiv.org/abs/2209.01540

[100] S. He *et al.*, "VLMAE: Vision-Language Masked Autoencoder." arXiv, Aug. 19, 2022. Accessed: Nov. 29, 2022. [Online]. Available: http://arxiv.org/abs/2208.09374

[101] Y. Gong *et al.*, "Contrastive Audio-Visual Masked Autoencoder." arXiv, Oct. 16, 2022. Accessed: Nov. 29, 2022. [Online]. Available: http://arxiv.org/abs/2210.07839

[102] P.-Y. Huang *et al.*, "Masked Autoencoders that Listen." arXiv, Jul. 26, 2022. Accessed: Nov. 30, 2022. [Online]. Available: http://arxiv.org/abs/2207.06405

[103] "Group masked autoencoder based density estimator for audio anomaly detection," *Amazon Science*. https://www.amazon.science/publications/group-masked-autoencoder-based-density-estimator-for-audio-anomaly-detection (accessed Nov. 30, 2022).

[104] A. Baade, P. Peng, and D. Harwath, "MAE-AST: Masked Autoencoding Audio Spectrogram Transformer." arXiv, Mar. 30, 2022. Accessed: Nov. 29, 2022. [Online]. Available: http://arxiv.org/abs/2203.16691

[105] D. Niizumi, D. Takeuchi, Y. Ohishi, N. Harada, and K. Kashino, "Masked Spectrogram Modeling using Masked Autoencoders for Learning General-purpose Audio Representation." arXiv, Apr. 26, 2022. Accessed: Nov. 29, 2022. [Online]. Available: http://arxiv.org/abs/2204.12260

[106] D. Niizumi, D. Takeuchi, Y. Ohishi, N. Harada, and K. Kashino, "Masked Modeling Duo: Learning Representations by Encouraging Both Networks to Model the Input." arXiv, Nov. 18, 2022. Accessed: Nov. 30, 2022. [Online]. Available: http://arxiv.org/abs/2210.14648

[107] E. Schwartz *et al.*, "MAEDAY: MAE for few and zero shot AnomalY-Detection." arXiv, Nov. 25, 2022. Accessed: Nov. 30, 2022. [Online]. Available: http://arxiv.org/abs/2211.14307

[108] N. Madan *et al.*, "Self-Supervised Masked Convolutional Transformer Block for Anomaly Detection." arXiv, Sep. 25, 2022. Accessed: Nov. 30, 2022. [Online]. Available: http://arxiv.org/abs/2209.12148

[109] H. Yao, X. Wang, and W. Yu, "Siamese Transition Masked Autoencoders as Uniform Unsupervised Visual Anomaly Detector." arXiv, Nov. 01, 2022. Accessed: Nov. 30, 2022. [Online]. Available: http://arxiv.org/abs/2211.00349

[110] Chuang Liu, Yibing Zhan, Xueqi Ma, Dapeng Tao, Bo Du, and Wenbin Hu, "Masked Graph Auto-Encoder Constrained Graph Pooling." [Online]. Available: https://2022.ecmlpkdd.org/wp-content/uploads/2022/09/sub_542.pdf

[111] S. Zhang, H. Chen, H. Yang, X. Sun, P. S. Yu, and G. Xu, "Graph Masked Autoencoders with Transformers." arXiv, May 12, 2022. Accessed: Nov. 30, 2022. [Online]. Available: http://arxiv.org/abs/2202.08391

[112] K. Jing, J. Xu, and P. Li, "Graph Masked Autoencoder Enhanced Predictor for Neural Architecture Search," presented at the Thirty-First International Joint Conference on Artificial Intelligence, Jul. 2022, vol. 4, pp. 3114–3120. doi: 10.24963/ijcai.2022/432.

[113] Z. Hou *et al.*, "GraphMAE: Self-Supervised Masked Graph Autoencoders." arXiv, Jul. 13, 2022. Accessed: Nov. 30, 2022. [Online]. Available: http://arxiv.org/abs/2205.10803

[114] Y. Tian, K. Dong, C. Zhang, C. Zhang, and N. V. Chawla, "Heterogeneous Graph Masked Autoencoders." arXiv, Aug. 21, 2022. Accessed: Nov. 30, 2022. [Online]. Available: http://arxiv.org/abs/2208.09957

[115] Q. Tan, N. Liu, X. Huang, R. Chen, S.-H. Choi, and X. Hu, "MGAE: Masked Autoencoders for Self-Supervised Learning on Graphs." arXiv, Jan. 07, 2022. Accessed: Nov. 30, 2022. [Online]. Available: http://arxiv.org/abs/2201.02534

[116] J. Li *et al.*, "MaskGAE: Masked Graph Modeling Meets Graph Autoencoders." arXiv, May 20, 2022. Accessed: Nov. 30, 2022. [Online]. Available: http://arxiv.org/abs/2205.10053

[117] X. Yu, L. Tang, Y. Rao, T. Huang, J. Zhou, and J. Lu, "Point-BERT: Pre-training 3D Point Cloud Transformers with Masked Point Modeling." arXiv, Jun. 06, 2022. Accessed: Nov. 29, 2022. [Online]. Available: http://arxiv.org/abs/2111.14819

[118] Y. Pang, W. Wang, F. E. H. Tay, W. Liu, Y. Tian, and L. Yuan, "Masked Autoencoders for Point Cloud Self-supervised Learning." arXiv, Mar. 28, 2022. Accessed: Nov. 29, 2022. [Online]. Available: http://arxiv.org/abs/2203.06604

[119] R. Zhang *et al.*, "Point-M2AE: Multi-scale Masked Autoencoders for Hierarchical Point Cloud Pre-training." arXiv, Oct. 13, 2022. Accessed: Nov. 29, 2022. [Online]. Available: http://arxiv.org/abs/2205.14401

[120] H. Liu, M. Cai, and Y. J. Lee, "Masked Discrimination for Self-Supervised Learning on Point Clouds." arXiv, Aug. 01, 2022. Accessed: Nov. 29, 2022. [Online]. Available: http://arxiv.org/abs/2203.11183

[121] G. Hess, J. Jaxing, E. Svensson, D. Hagerman, C. Petersson, and L. Svensson, "Masked Autoencoder for Self-Supervised Pre-training on Lidar Point Clouds." arXiv, Oct. 24, 2022. Accessed: Nov. 30, 2022. [Online]. Available: http://arxiv.org/abs/2207.00531

[122] C. Min, X. Xu, D. Zhao, L. Xiao, Y. Nie, and B. Dai, "Voxel-MAE: Masked Autoencoders for Self-supervised Pre-training Large-scale Point Clouds." arXiv, Nov. 23, 2022. Accessed: Nov. 29, 2022. [Online]. Available: http://arxiv.org/abs/2206.09900

[123] H. Rao and C. Miao, "SimMC: Simple Masked Contrastive Learning of Skeleton Representations for Unsupervised Person Re-Identification," Apr. 2022, doi: 10.48550/arXiv.2204.09826.

[124] S. Y. Kim *et al.*, "Layered Depth Refinement with Mask Guidance." arXiv, Jun. 07, 2022. Accessed: Nov. 30, 2022. [Online]. Available: http://arxiv.org/abs/2206.03048

[125] Y. Wang, Z. Pan, X. Li, Z. Cao, K. Xian, and J. Zhang, "Less is More: Consistent Video Depth Estimation with Masked Frames Modeling," in *Proceedings of the 30th ACM International Conference on Multimedia*, Oct. 2022, pp. 6347–6358. doi: 10.1145/3503161.3547978.

[126] T. Yu, Z. Zhang, C. Lan, Y. Lu, and Z. Chen, "Mask-based Latent Reconstruction for Reinforcement Learning." arXiv, Oct. 09, 2022. Accessed: Nov. 30, 2022. [Online]. Available: http://arxiv.org/abs/2201.12096

[127] Y. Seo *et al.*, "Masked World Models for Visual Control." arXiv, Nov. 15, 2022. Accessed: Nov. 30, 2022. [Online]. Available: http://arxiv.org/abs/2206.14244

[128] Y. Liang, S. Zhao, B. Yu, J. Zhang, and F. He, "MeshMAE: Masked Autoencoders for 3D Mesh Data Analysis." arXiv, Jul. 20, 2022. Accessed: Nov. 30, 2022. [Online]. Available: http://arxiv.org/abs/2207.10228

[129] W. Xu, C. Zhang, F. Zhao, and L. Fang, "A Mask-Based Adversarial Defense Scheme," Apr. 2022, doi: 10.48550/arXiv.2204.11837.

[130] J. Jiang, J. Chen, and Y. Guo, "A Dual-Masked Auto-Encoder for Robust Motion Capture with Spatial-Temporal Skeletal Token Completion." arXiv, Jul. 15, 2022. Accessed: Nov. 30, 2022. [Online]. Available: http://arxiv.org/abs/2207.07381

[131] H.-Y. S. Chien, H. Goh, C. M. Sandino, and J. Y. Cheng, "MAEEG: Masked Auto-encoder for EEG Representation Learning." arXiv, Oct. 27, 2022. Accessed: Nov. 30, 2022. [Online]. Available: http://arxiv.org/abs/2211.02625

[132] Z. Yang, Z. Li, M. Shao, D. Shi, Z. Yuan, and C. Yuan, "Masked Generative Distillation." arXiv, Jul. 05, 2022. Accessed: Nov. 30, 2022. [Online]. Available: http://arxiv.org/abs/2205.01529

[133] M. Germain, K. Gregor, I. Murray, and H. Larochelle, "MADE: Masked Autoencoder for Distribution Estimation." arXiv, Jun. 05, 2015. Accessed: Nov. 29, 2022. [Online]. Available: http://arxiv.org/abs/1502.03509

[134] F. Liu, H. Liu, A. Grover, and P. Abbeel, "Masked Autoencoding for Scalable and Generalizable Decision Making." arXiv, Nov. 23, 2022. Accessed: Nov. 30, 2022. [Online]. Available: http://arxiv.org/abs/2211.12740

[135] G. Kutiel, R. Cohen, M. Elad, and D. Freedman, "What's Behind the Mask: Estimating Uncertainty in Image-to-Image Problems." arXiv, Nov. 28, 2022. Accessed: Nov. 30, 2022. [Online]. Available: http://arxiv.org/abs/2211.15211

[136] H. Guo, H. Zhu, J. Wang, V. Prahlad, W. K. Ho, and T. H. Lee, "Masked Self-Supervision for Remaining Useful Lifetime Prediction in Machine Tools." arXiv, Jul. 04, 2022. Accessed: Nov. 30, 2022. [Online]. Available: http://arxiv.org/abs/2207.01219

[137] K. Majmundar, S. Goyal, P. Netrapalli, and P. Jain, "MET: Masked Encoding for Tabular Data." arXiv, Jun. 17, 2022. Accessed: Nov. 30, 2022. [Online]. Available: http://arxiv.org/abs/2206.08564

[138] Y. Hao, R. Wang, Z. Cao, Z. Wang, Y. Cui, and D. Sadigh, "Masked Imitation Learning: Discovering Environment-Invariant Modalities in Multimodal Demonstrations." arXiv, Sep. 15, 2022. Accessed: Nov. 30, 2022. [Online]. Available: http://arxiv.org/abs/2209.07682

[139] I. Radosavovic, T. Xiao, S. James, P. Abbeel, J. Malik, and T. Darrell, "Real-World Robot Learning with Masked Visual Pre-training." arXiv, Oct. 06, 2022. Accessed: Nov. 30, 2022. [Online]. Available: http://arxiv.org/abs/2210.03109

[140] M. Zha, S. Wong, M. Liu, T. Zhang, and K. Chen, "Time Series Generation with Masked Autoencoder." arXiv, May 19, 2022. Accessed: Nov. 30, 2022. [Online]. Available: http://arxiv.org/abs/2201.07006

[141] C. Zhang, C. Zhang, J. Song, J. S. K. Yi, K. Zhang, and I. S. Kweon, "A Survey on Masked Autoencoder for Self-supervised Learning in Vision and Beyond." arXiv, Jul. 30, 2022. Accessed: Nov. 30, 2022. [Online]. Available: http://arxiv.org/abs/2208.00173

[142] Y. Wei *et al.*, "Contrastive Learning Rivals Masked Image Modeling in Fine-tuning via Feature Distillation," May 2022, doi: 10.48550/arXiv.2205.14141.

[143] Z. Fu, W. Zhou, J. Xu, H. Zhou, and L. Li, "Contextual Representation Learning beyond Masked Language Modeling." arXiv, Apr. 08, 2022. Accessed: Nov. 29, 2022. [Online]. Available: http://arxiv.org/abs/2204.04163

[144] Z. Xie *et al.*, "On Data Scaling in Masked Image Modeling." arXiv, Jun. 09, 2022. doi: 10.48550/arXiv.2206.04664.

[145] Y. Fang *et al.*, "EVA: Exploring the Limits of Masked Visual Representation Learning at Scale." arXiv, Nov. 14, 2022. Accessed: Nov. 30, 2022. [Online]. Available: http://arxiv.org/abs/2211.07636

[146] K. Zhang and Z. Shen, "i-MAE: Are Latent Representations in Masked Autoencoders Linearly Separable?" arXiv, Oct. 20, 2022. Accessed: Nov. 30, 2022. [Online]. Available: http://arxiv.org/abs/2210.11470

[147] Z. Xie, Z. Geng, J. Hu, Z. Zhang, H. Hu, and Y. Cao, "Revealing the Dark Secrets of Masked Image Modeling," May 2022, doi: 10.48550/arXiv.2205.13543.

[148] S. Cao, P. Xu, and D. A. Clifton, "How to Understand Masked Autoencoders." arXiv, Feb. 09, 2022. Accessed: Nov. 30, 2022. [Online]. Available: http://arxiv.org/abs/2202.03670

[149] X. Kong and X. Zhang, "Understanding Masked Image Modeling via Learning Occlusion Invariant Feature." arXiv, Aug. 08, 2022. Accessed: Nov. 30, 2022. [Online]. Available: http://arxiv.org/abs/2208.04164

[150] J. Pan, P. Zhou, and S. Yan, "Towards Understanding Why Mask-Reconstruction Pretraining Helps in Downstream Tasks." arXiv, Jun. 14, 2022. Accessed: Nov. 30, 2022. [Online]. Available: http://arxiv.org/abs/2206.03826

[151] F. Xue *et al.*, "Deeper vs Wider: A Revisit of Transformer Configuration," May 2022, doi: 10.48550/arXiv.2205.10505.

[152] I. Kong, D. Yang, J. Lee, I. Ohn, and Y. Kim, "Masked Bayesian Neural Networks : Computation and Optimality." arXiv, Jun. 01, 2022. Accessed: Nov. 30, 2022. [Online]. Available: http://arxiv.org/abs/2206.00853

[153] S. Li *et al.*, "Architecture-Agnostic Masked Image Modeling -- From ViT back to CNN," May 2022, doi: 10.48550/arXiv.2205.13943.

[154] S. Woo *et al.*, "ConvNeXt V2: Co-designing and Scaling ConvNets with Masked Autoencoders." arXiv, Jan. 02, 2023. Accessed: Jan. 10, 2023. [Online]. Available: http://arxiv.org/abs/2301.00808

[155] L. Liu *et al.*, "Deep Learning for Generic Object Detection: A Survey." arXiv, Aug. 22, 2019. Accessed: Dec. 21, 2022. [Online]. Available: http://arxiv.org/abs/1809.02165

[156] "ResNeXt: Aggregated Residual Transformations for Deep Neural Networks." Meta Research, Jan. 04, 2023. Accessed: Jan. 09, 2023. [Online]. Available: https://github.com/facebookresearch/ResNeXt

[157] "ImageNet." https://www.image-net.org/ (accessed Nov. 30, 2022).

[158] J. Devlin, M.-W. Chang, K. Lee, and K. Toutanova, "BERT: Pre-training of Deep Bidirectional Transformers for Language Understanding." arXiv, May 24, 2019. Accessed: Nov. 30, 2022. [Online]. Available: http://arxiv.org/abs/1810.04805

[159] "Generative_Pretraining_from_Pixels_V2.pdf." Accessed: Dec. 02, 2022. [Online]. Available: https://cdn.openai.com/papers/Generative_Pretraining_from_Pixels_V2.pdf

[160] X. Xie, G. Cheng, J. Wang, X. Yao, and J. Han, "Oriented R-CNN for Object Detection." arXiv, Aug. 12, 2021. Accessed: Dec. 04, 2022. [Online]. Available: http://arxiv.org/abs/2108.05699

[161] J. Han, J. Ding, N. Xue, and G.-S. Xia, "ReDet: A Rotation-equivariant Detector for Aerial Object Detection." arXiv, Mar. 13, 2021. Accessed: Dec. 03, 2022. [Online]. Available: http://arxiv.org/abs/2103.07733

[162] R. Girshick, J. Donahue, T. Darrell, and J. Malik, "Rich feature hierarchies for accurate object detection and semantic segmentation." arXiv, Oct. 22, 2014. Accessed: Dec. 03, 2022. [Online]. Available: http://arxiv.org/abs/1311.2524

[163] S. Ren, K. He, R. Girshick, and J. Sun, "Faster R-CNN: Towards Real-Time Object Detection with Region Proposal Networks." arXiv, Jan. 06, 2016. Accessed: Dec. 03, 2022. [Online]. Available: http://arxiv.org/abs/1506.01497

[164] R. Girshick, "Fast R-CNN." arXiv, Sep. 27, 2015. Accessed: Jan. 03, 2023. [Online]. Available: http://arxiv.org/abs/1504.08083

[165] "Selective Search for Object Detection | R-CNN," *GeeksforGeeks*, Feb. 25, 2020. https://www.geeksforgeeks.org/selective-search-for-object-detection-r-cnn/ (accessed Jan. 03, 2023).

[166] K. He, G. Gkioxari, P. Dollár, and R. Girshick, "Mask R-CNN." arXiv, Jan. 24, 2018. Accessed: Nov. 30, 2022. [Online]. Available: http://arxiv.org/abs/1703.06870

[167] J. Long, E. Shelhamer, and T. Darrell, "Fully Convolutional Networks for Semantic Segmentation." arXiv, Mar. 08, 2015. Accessed: Dec. 03, 2022. [Online]. Available: http://arxiv.org/abs/1411.4038

[168] Z. Cai and N. Vasconcelos, "Cascade R-CNN: Delving into High Quality Object Detection." arXiv, Dec. 03, 2017. Accessed: Dec. 03, 2022. [Online]. Available: http://arxiv.org/abs/1712.00726

[169] Y. Zhou *et al.*, "MMRotate: A Rotated Object Detection Benchmark using PyTorch," in *Proceedings of the 30th ACM International Conference on Multimedia*, Oct. 2022, pp. 7331–7334. doi: 10.1145/3503161.3548541.

[170] A. Vaswani *et al.*, "Attention Is All You Need." arXiv, Dec. 05, 2017. Accessed: Dec. 21, 2022. [Online]. Available: http://arxiv.org/abs/1706.03762

[171] J. Shermeyer, T. Hossler, A. Van Etten, D. Hogan, R. Lewis, and D. Kim, "RarePlanes: Synthetic Data Takes Flight." arXiv, Nov. 10, 2020. Accessed: Dec. 04, 2022. [Online]. Available: http://arxiv.org/abs/2006.02963

[172] "DOTA." https://captain-whu.github.io/DOTA/evaluation.html (accessed Dec. 04, 2022).

[173] "Airbus Aircraft Detection." https://www.kaggle.com/datasets/airbusgeo/airbus-aircrafts-sample-dataset (accessed Dec. 04, 2022).

[174] G.-S. Xia *et al.*, "DOTA: A Large-scale Dataset for Object Detection in Aerial Images," Nov. 2017, doi: 10.48550/arXiv.1711.10398.

[175] "facebookresearch/detectron2." Meta Research, Dec. 04, 2022. Accessed: Dec. 04, 2022. [Online]. Available: https://github.com/facebookresearch/detectron2

[176] "OpenMMLab." https://github.com/open-mmlab (accessed Dec. 04, 2022).

[177] MMDetection Contributors, "OpenMMLab Detection Toolbox and Benchmark." Aug. 2018. Accessed: Dec. 21, 2022. [Online]. Available: https://github.com/open-mmlab/mmdetection

[178] "facebookresearch/mae." Meta Research, Jan. 10, 2023. Accessed: Jan. 10, 2023. [Online]. Available: https://github.com/facebookresearch/mae

[179] X. Yang *et al.*, "Learning High-Precision Bounding Box for Rotated Object Detection via Kullback-Leibler Divergence." arXiv, Apr. 18, 2022. Accessed: Jan. 04, 2023. [Online]. Available: http://arxiv.org/abs/2106.01883